\pdfoutput=1

\documentclass[11pt]{article}

\usepackage[final]{acl}

\usepackage{times}
\usepackage{latexsym}

\usepackage[T1]{fontenc}

\usepackage[utf8]{inputenc}

\usepackage{microtype}

\usepackage{inconsolata}

\usepackage{graphicx}

\usepackage{array}
\usepackage{listings}
\usepackage[table]{xcolor}
\usepackage[most]{tcolorbox}

\usepackage{lipsum}
\usepackage{geometry}

\usepackage{microtype}
\usepackage{url}
\usepackage{booktabs}
\usepackage{tabularx}

\usepackage{subfigure}
\usepackage{enumitem}
\newenvironment{itemize*}%
 {\leftmargini=20pt\begin{itemize}%
  \setlength{\itemsep}{3pt}%
  \setlength{\parskip}{0pt}%
  }%
 {\end{itemize}}
\newenvironment{enumerate*}%
 {\begin{enumerate}%
  \setlength{\itemsep}{0pt}%
  \setlength{\parskip}{0pt}}%
 {\end{enumerate}}

\usepackage{amsmath}
\usepackage{cleveref}
\usepackage{listings}
\usepackage{titlesec}
\usepackage{multirow}
\usepackage{makecell}
\usepackage{wrapfig}
\usepackage{booktabs}
\usepackage{makecell}

\usepackage{array}       
\usepackage{caption}     
\usepackage{amssymb}
\usepackage{amsthm}
\usepackage{dsfont}
\usepackage{amsmath}          
\usepackage{booktabs}         
\usepackage{tabularx}         
\usepackage{graphicx}         
\usepackage[utf8]{inputenc}  
\usepackage{hyperref}
\usepackage{cuted}

\definecolor{midnightgreen}{rgb}{0.0, 0.29, 0.33}
\definecolor{deepgreen}{HTML}{0aa344}
\definecolor{deeppurple}{HTML}{7030a0}
\definecolor{deepblue}{HTML}{171d91}
\definecolor{brown}{HTML}{843c0c}
\definecolor{shadered}{HTML}{ffe5e5}
\definecolor{shadegreen}{HTML}{e5f7ed}
\definecolor{msftBlack}{RGB}{0,0,0}
\definecolor{lightred}{RGB}{255,163,163}
\definecolor{deepred}{RGB}{200,0,0}

\usepackage{xcolor}
\definecolor{grey}{RGB}{128,128,128}

\usepackage{pifont}
\newcommand{\cmark}{\textcolor[rgb]{0.0, 0.6, 0.0}{\ding{51}}} 
\newcommand{\xmark}{\textcolor[rgb]{0.7, 0.0, 0.0}{\ding{55}}} 
\newcommand{\gmark}{\textcolor[rgb]{1,0.647,0}{\ding{51}}}

\NewDocumentCommand{\jiayu}
{ mO{} }{\textcolor{blue}{\textsuperscript{\textit{Jiayu}}\textsf{\textbf{\small[#1]}}}}

\NewDocumentCommand{\cheng}
{ mO{} }{\textcolor{orange}{\textsuperscript{\textit{Cheng}}\textsf{\textbf{\small[#1]}}}}

\NewDocumentCommand{\qzong}
{ mO{} }{\textcolor{blue}{\textsuperscript{\textit{Qing}}\textsf{\textbf{\small[#1]}}}}

\NewDocumentCommand{\yi}
{ mO{} }{\textcolor{teal}{\textsuperscript{\textit{Yi}}\textsf{\textbf{\small[#1]}}}}

\newcommand{\CB}[1]{\textbf{CostBench}}
\newcommand{\cb}[1]{CostBench}
    
\title{CostBench: Evaluating Multi-Turn Cost-Optimal Planning and Adaptation in Dynamic Environments for LLM Tool-Use Agents}

\author{
\textbf{Jiayu Liu$^{1}$ ~~Cheng Qian$^{2}$ ~~Zhaochen Su$^{1}$ ~~Qing Zong$^{1}$ ~~Shijue Huang$^{1}$ ~~Bingxiang He$^{3}$} \\ 
\textbf{Yi R. (May) Fung$^{1\dagger}$}\\
$^{1}$The Hong Kong University of Science and Technology \\
$^{2}$University of Illinois Urbana-Champaign ~~~$^{3}$Tsinghua University \\
\texttt{jliufv@connect.ust.hk}
~~~\texttt{yrfung@ust.hk}
}

\begin{document}
\maketitle

\vspace{-1.2cm}
\begin{strip}
\vspace{-1.6cm}
\centering
\small
\newcommand{\logoh}{1.35em}

\href{https://github.com/JiayuJeff/CostBench}{%
\raisebox{-0.2\height}{\includegraphics[height=\logoh]{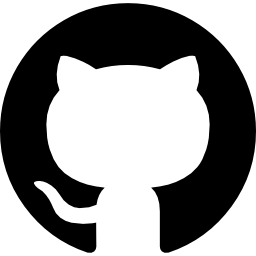}}%
\hspace{0.35em}\textbf{Code}%
}
\quad
\href{https://huggingface.co/datasets/JiayuJeff/CostBench}{%
\raisebox{-0.2\height}{\includegraphics[height=\logoh]{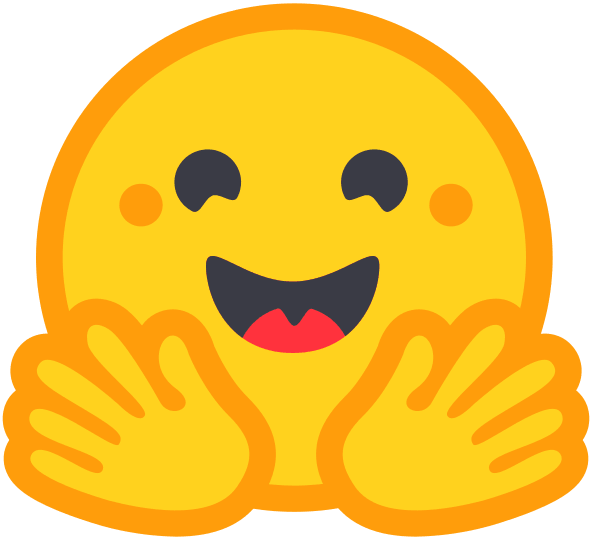}}%
\hspace{0.35em}\textbf{Dataset}%
}
\end{strip}

\begin{abstract}

Current evaluations of Large Language Model (LLM) agents primarily emphasize task completion, often overlooking resource efficiency and adaptability. 
This neglects a crucial capability: agents’ ability to devise and adjust cost-optimal plans in response to changing environments. 
To bridge this gap, we introduce \CB{}, a scalable, cost-centric benchmark designed to evaluate agents’ economic reasoning and replanning abilities. 
Situated in the travel-planning domain, \cb{} comprises tasks solvable via multiple sequences of atomic and composite tools with diverse, customizable costs. 
It also supports four types of dynamic blocking events, such as tool failures and cost changes, to simulate real-world unpredictability and require agents to adapt in real time.
Evaluating leading open-sourced and proprietary models on \cb{} reveals a substantial gap in cost-aware planning: 
agents frequently fail to identify cost-optimal solutions in static settings, with even \textit{GPT-5} achieving less than 75\% exact match rate on the hardest tasks, and performance further drops significantly under dynamic conditions.
By diagnosing these weaknesses, \cb{} lays the groundwork for developing future agents that are both economically rational and robust.\footnote{$\dagger$ denotes corresponding author.}



\end{abstract}

\section{Introduction}\label{sec:intro}
\vspace{-0.7em}
Large Language Models (LLMs) have demonstrated remarkable capabilities in reasoning~\cite{deepseek-r1, Multirole-R1}, code generation~\cite{code-good-1,code-good-2}, and complex problem-solving~\cite{DAPO,complex-reasoning-good-1}.
When equipped with external tools, they can interact with dynamic environments such as the web~\cite{WebResearcher,he2025advancing} and interactive systems~\cite{qin2024toollearningfoundationmodels, su2025openthinkimg}.
This integration of tool use gives LLMs agentic capabilities, enabling them to autonomously execute multi-step tasks.

While tool-augmented reasoning provides impressive autonomy, its effectiveness ultimately hinges on strategic \textit{planning}: deciding not only which tools to use but also how to sequence them efficiently to achieve complex objectives.
Existing evaluations, however, focus primarily on task completion~\cite{ToolQA, MINT, ACE-Bench}, offering limited insight into how different tool-use strategies affect resource consumption.
Although prior studies have examined factors such as API fees~\cite{Cost-API-1, Cost-API-2}, token usage~\cite{Cost-token-1, AdaCtrl}, GPU computation~\cite{Cost-GPU-1}, and memory overhead~\cite{CATP-LLM}, they often overlook the agent’s ability to reason about costs and remain cost-aware in dynamic environments.
As a result, it remains unclear whether current agents can effectively plan and adapt under shifting cost conditions.
This gap motivates our central question:
\textbf{how effectively can LLM agents devise and adapt cost-optimal plans for arbitrary cost functions in dynamic environments?}
Answering this requires an evaluation environment that ~\textit{i)} features \textbf{diverse and flexible cost structures} to reveal the agent’s cost sensitivity, and ~\textit{ii)} introduces \textbf{runtime dynamics} that evolve throughout the interaction, challenging the agent to sustain cost-awareness and replan adaptively.

\begin{table*}[!t]
\vspace{-0.1in}
\centering
\resizebox{\linewidth}{!}{
    \begin{tabular}{lcccccccccc}
    \toprule
    \textbf{Benchmark} 
    & \textbf{\makecell{Multi-turn\\Interaction}} 
    & \textbf{\makecell{Cost-optimal\\Planning}} 
    & \textbf{\makecell{Flexible\\Cost}} 
    & \textbf{\makecell{Tool\\Use}} 
    & \textbf{\makecell{Dynamic\\State}} 
    & \textbf{\makecell{Adjustable\\Difficulty}} 
    & \textbf{\makecell{Customizable\\Details}} 
    & \textbf{\makecell{Scal-\\able}}\\
    \midrule
    \textit{PlanBench~\citep{PlanBench}} 
    &  \xmark 
    &  \cmark
    &  \cmark 
    &  \xmark 
    &  \xmark 
    &  \cmark
    &  \cmark
    &  \cmark 
    \\
    \textit{ToolBench~\citep{ToolBench}}          
    & \cmark
    & \gmark 
    & \xmark 
    & \cmark
    & \xmark 
    & \cmark
    & \cmark
    & \cmark
    \\
    \textit{TravelPlanner~\citep{TravelPlanner}}          
    & \cmark
    & \xmark 
    & \xmark 
    & \cmark
    & \gmark 
    & \cmark
    & \xmark 
    & \gmark 
    \\
    \textit{AucArena~\citep{AucArena}}                
    & \cmark
    & \xmark 
    & \xmark 
    & \xmark 
    & \cmark 
    & \cmark 
    & \cmark
    & \gmark
    \\
    \textit{SayCanPay~\citep{SayCanPay}}              
    & \xmark
    & \cmark
    & \xmark
    & \xmark
    & \xmark
    & \cmark
    & \cmark
    & \cmark
    \\
    \textit{ACPBench Hard~\citep{ACPBench-Hard}}                           
    & \xmark
    & \cmark
    & \xmark
    & \xmark
    & \xmark
    & \xmark 
    & \cmark
    & \cmark
    \\
    \textit{UserBench~\citep{UserBench}}  
    & \cmark
    & \gmark 
    & \xmark
    & \cmark
    & \cmark
    & \cmark
    & \cmark
    & \cmark
    \\
    
    \textit{MCP-Bench~\citep{MCP-Bench}} 
    & \cmark
    & \xmark 
    & \xmark 
    & \cmark
    & \xmark 
    & \cmark
    & \xmark 
    & \cmark
    \\
    \textit{MINT~\citep{MINT}} 
    & \cmark
    & \xmark 
    & \xmark 
    & \cmark
    & \xmark 
    & \cmark
    & \gmark 
    & \gmark
    \\
    \midrule
    \textbf{\CB}~\textbf{(Ours)}              
    & \cmark  
    & \cmark  
    & \cmark  
    & \cmark  
    & \cmark  
    & \cmark  
    & \cmark  
    & \cmark 
    \\
    \bottomrule
    \end{tabular}
}
\caption{For each benchmark, the table reports whether each trait is fully (\cmark), partially (\gmark), or not (\xmark) addressed. Detailed explanations are provided in Appendix~\ref{app:Comparison-of-related-work}. 
}
\label{tab:comparison-table}
\vspace{-0.25in}
\end{table*}

To this end, we introduce \CB{}, a scalable and cost-centric benchmark situated in the travel-planning domain. 
\cb{} defines six travel-related tasks, each of which can be completed through sequences of non-separable atomic tools. 
The costs of these atomic tools are randomly assigned, enabling diverse cost configurations across different runs.
Building on atomic tools, we create composite ones whose costs equal the aggregated component costs plus Gaussian noise. 
This design creates multiple alternative pathways to accomplish the same goal, enriching the benchmark’s dynamics and encouraging agents to maintain cost-awareness when selecting the optimal execution path. 
\cb{} also introduces dynamic blocking events (e.g., tool cost or user preference changes) that simulate four types of real-world disruptions during multi-turn interactions, requiring agents to replan their trajectories while maintaining cost-optimality.

We evaluate ten leading open-source and proprietary LLMs on \cb{} and find that all models perform poorly. \textit{GPT-5} yields the best performance yet attains less than 75\% exact match rate (i.e., the tool-call sequence exactly matches the ground-truth trajectory)
in the most difficult static settings, and its accuracy further declines to approximately 35\% under cost-change conditions. 
Overall, the models are highly sensitive to cost noise and environmental perturbations, especially when tool costs fluctuate or tools become corrupted after use.
These results highlight the fragility of current LLM agents in maintaining cost-awareness and the necessity for more robust, dynamically adaptive models.
We summarize our main contributions as follows:
\begin{itemize}[topsep=0pt, partopsep=1.5pt, leftmargin=*, itemsep=-3pt]
    \item \textbf{Scalable Benchmark Framework:} We introduce CostBench, a scalable and cost-centric framework built on a task-generation pipeline. It programmatically generates tasks with multiple atomic and composite tools, featuring diverse and tunable cost assignments to probe agents' economic reasoning.
    
    \item \textbf{Dynamic Interaction Environment:} We design an interactive, multi-turn environment that simulates real-world unpredictability. It introduces four types of dynamic blocking events (e.g., cost changes, tool failures) to compel agents to adapt and replan in real time. The environment also serves as a reinforcement learning (RL) playground for agent refinement.
    
    \item \textbf{Comprehensive Analysis:} We conduct a comprehensive evaluation of ten leading LLMs, revealing their high sensitivity to cost variations and dynamic disruptions. Our analysis highlights a substantial gap in cost-aware planning, with performance dropping by around 40\% under dynamic conditions, underscoring limitations in achieving robust, cost-optimal planning.
\end{itemize}
Looking ahead, \CB{} lays the foundation for future research on adaptive and cost-aware LLM agents, encouraging the development of models that can dynamically plan, learn, and optimize under evolving real-world constraints. 
\vspace{-0.4em}

\section{Related Work}
\vspace{-0.3em}
\begin{figure*}[ht]
    \centering
    \includegraphics[width=\linewidth]{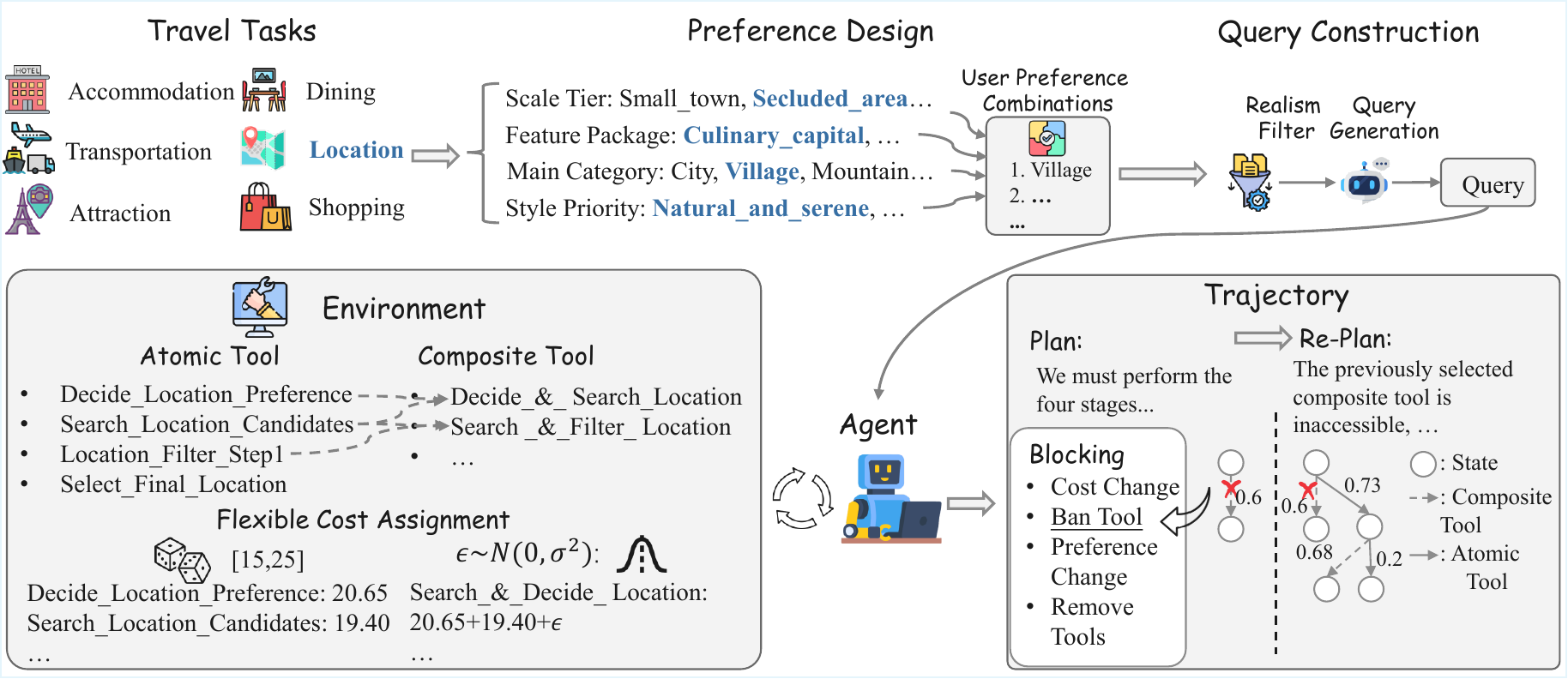}
    \caption{Overview of the \CB{} pipeline.
    Starting from high-quality queries generated from combinations of user preferences, the agent constructs its plan, then interacts with an environment set up with atomic and composite tools under flexible cost assignments (atomic tool costs are randomized between 15 and 25 in our experiments), and executes actions along a customizable dynamic blocking module to achieve its goal.}
    \label{fig:overview-pipeline}
    \vspace{-0.2in}
\end{figure*}

\paragraph{Evaluation of LLM cost-optimal planning.}

Recent work emphasizes evaluating LLMs under economically grounded conditions. Existing tool-use benchmarks such as MINT~\cite{MINT}, ToolQA~\cite{ToolQA}, and MTU-Bench~\cite{MTU-Bench} focus on task completion but ignore efficiency. Cost-related benchmarks like PlanBench~\cite{PlanBench}, SayCanPay~\cite{SayCanPay}, and ACPBench Hard~\cite{ACPBench-Hard, ACP-Bench} consider cost only in single-turn planning, limiting applicability to dynamic, iterative settings. Multi-turn benchmarks such as UserBench~\cite{UserBench} and TravelPlanner~\cite{TravelPlanner} focus on candidate prices rather than operational costs, while others like R-ConstraintBench~\cite{R-ConstraintBench} and Flex-TravelPlanner~\cite{flex-TravelPlanner} emphasize performance–resource trade-offs under fixed budgets. Most also conflate cost with path functionality and use static environments, obscuring cost sensitivity. In contrast, \cb{} decouples cost from functional bias through randomized tool assignments and equivalent solution paths, while introducing dynamic blocking events to assess adaptive planning. A detailed comparison with prior work is provided in Table~\ref{tab:comparison-table}.



\paragraph{Cost-centric Agent Design.}
Prior work has examined cost-awareness in agents from several angles. Number of tool call represents the simplest form of tool cost which is thoroughly examined and optimized through algorithms~\citep{qian2025smart, wang2025acting, wang2025toward}. 
Furthermore, Toolformer~\cite{Toolformer} focuses on when to invoke tools rather than reasoning about fine-grained tool costs or long-term expenditure. 
Starting from FrugalGPT~\cite{FrugalGPT}, many studies further explore multi-agent coordination to reduce expenses, achieving cost control through selective and hierarchical model invocation~\cite{EcoAssistant,Adaptive-LLM-Router,BudgetMLAgent}, often in limited or static scenarios. SayCanPay~\cite{SayCanPay} and CATP-LLM~\cite{CATP-LLM} explicitly optimize for cost but remain limited to one-off planning without dynamic adaptation to changing environments. Other studies treat cost merely as token usage and focus on efficient reasoning under token constraints~\cite{token-budget-1,token-budget-2,AdaCtrl}, neglecting broader operational costs. Overall, these works target narrow aspects of cost optimization, whereas our benchmark significantly broadens the notion of cost and systematically evaluates agent adaptability under realistic, dynamically changing conditions and diverse practical constraints.

\section{\CB{}}

\CB{} is an interactive, consumption-aware environment designed to evaluate agents’ ability for cost-optimal planning in complex reasoning tasks, where the goal is to complete a task while minimizing total cost.
It simulates realistic settings in which every action incurs a measurable cost.
To enable this, \cb{} offers a flexible ecosystem of tools encompassing both atomic operations and higher-level composite functions.
Each tool is assigned a randomized cost profile independently resampled for every data instance, preventing memorization and data leakage while encouraging agents to adopt adaptive, cost-sensitive strategies.

Moreover, \cb{} supports evaluation in two modes: a \textbf{static mode}, where user preferences, tool costs, and availability remain constant; and a \textbf{dynamic mode}, where these attributes evolve with agent behavior or environmental changes, capturing the uncertainty in real-world scenarios.
Each task unfolds through an iterative interaction loop in which the agent observes the task state, reasons about its next action, invokes relevant tools, and receives cost-tagged feedback that updates its internal state.
This continual feedback enables adaptive decision-making and dynamic replanning until the task goal is achieved.
We describe the detailed design below and discuss the significance and generalization of \cb{} in Section~\ref{sec:discussion}.




\subsection{Environment Setup}
In this work, we focus on the travel planning domain.
Following TravelPlanner~\cite{TravelPlanner} and UserBench~\cite{UserBench}, we categorize travel-related activities into six domains: \textit{Location}, \textit{Transportation}, \textit{Accommodation}, \textit{Attraction}, \textit{Dining}, and \textit{Shopping}.
Each task in \cb{} follows a structured four-stage workflow comprising \textit{preference identification}, \textit{candidate search}, \textit{candidate filtering}, and \textit{final selection}.
Among these, the filtering stage involves multiple steps, while the other three are modeled as single-step operations.
We define the task sequence as the total number of steps across all stages, providing a clear metric for task length and complexity.
This multi-step design enhances the scalability of the benchmark, enabling the flexible generation of tasks with varying lengths and levels of difficulty. \vspace{-0.1in}

\begin{table}[t]
    \centering
    \scriptsize
    \begin{tabular}{@{} l c c c @{}} 
        \toprule
        \textbf{Task Name} & 
        \makecell{\textbf{Pref. Comb.} \\ \textbf{(Unfiltered)} }
        &
        
        \makecell{\textbf{Train Split} \\ \textbf{Filtered} } & 
        
        \makecell{\textbf{Test Split} \\ \textbf{Filtered} }  \\
        \midrule
        Location  & 1552&  127 \phantom{0}(6.68\%) & \phantom{0}29 \phantom{0}(7.61\%)\\
        Transportation  & 1552  &  296 (15.56\%) & 102 (26.77\%)\\
        Accommodation  & 1552  &  417 (21.92\%) & \phantom{0}51 (13.39\%) \\
        Attraction  & 1552  &  361 (18.98\%) & \phantom{0}56 (14.70\%)\\
        Dining  & 1552  &  282 (14.83\%) & \phantom{0}60 (15.75\%)\\
        Shopping  & 1552  &  419 (22.03\%) & \phantom{0}83 (21.78\%) \\ 
        \midrule
        \textbf{Total / Proportion.} & \textbf{9312}& \textbf{1902 (100\%)} & \textbf{381 (100\%)}  \\ 
        \bottomrule
    \end{tabular}
    \vspace{-0.05in}

    \par
    \caption{Main statistics of query construction in \cb{}. We generate 6,000 training and 1,000 test user preference combinations and filter out those violating common sense. The total number of user preference combinations equals the sum of both splits, with a discussion of the slightly imbalanced distribution provided in Appendix~\ref{app:data-distribution}.}

    \label{tab:benchmark-statistics_query}
    \vspace{-0.2in}
\end{table}

\paragraph{Tool Library.}
To enable precise progress tracking, we introduce a set of self-defined data types that explicitly specify and strictly enforce the input–output schema for each tool.
This design prevents agents from skipping intermediate tool calling steps through direct guessing and allows the environment to automatically verify whether each action meaningfully contributes to the final solution.
In \cb{}, each task step maps to an \textit{atomic tool} performing a non-separable operation, while a \textit{composite tool} is a higher-level tool executing a sequence of atomic tools. For example, ``Search\_and\_Decide\_Location'' combines ``Decide\_Location\_Preference'' and ``Search\_Location\_Preference'' into an independent tool equivalent to running them sequentially (See Figure~\ref{fig:overview-pipeline}).
All tools follow the sequential order of task steps, with each tool consuming the outputs of preceding steps as its inputs (See example in Figure~\ref{fig:tool-flow-example}). 
This sequential dependency enables any consecutive subset of atomic tools to form a valid composite tool, thereby supporting scalability and the construction of longer or more complex tool chains while maintaining explicit input–output relationships.
Details of tool construction and an example of the tool schema are illustrated in Appendix~\ref{app:tool-construction-details} and Figure~\ref{fig:sample-tools}.

\paragraph{Tool Cost Assignment.}
Each atomic tool is assigned a randomized cost drawn from a customizable range, while the cost of a composite tool is defined as the sum of its constituent atomic tools plus Gaussian noise with zero mean and a  customizable standard deviation.
Tool costs are independently resampled for every data instance but remain reproducible, ensuring environment variations are diverse yet comparable.
For tasks sharing the same sequence length, cost assignments are kept consistent across static and dynamic blocking settings to ensure fair comparison.
Further implementation details are provided in Appendix~\ref{app:cost-assignment}.

\paragraph{Agent State Definition.}
At each time step, the agent observes the current state $s_t = {q, \mathcal{T}, \mathcal{C}, \mathcal{D}}$, where $q$ denotes the user query, $\mathcal{T}$ represents the available tool library, $\mathcal{C}$ specifies the corresponding tool cost table, and $\mathcal{D}$ contains the set of self-defined data types obtained from previous tool invocations.
Once a data type is acquired, it remains accessible throughout the interaction, reflecting realistic scenarios in which previously gathered information can be reused.
The agent’s objective is to reach the designated goal state specified clearly in its input while minimizing the total accumulated cost over the entire interaction trajectory.

\subsection{Agent–Environment Interaction}

Each task instance in \cb{} follows an iterative interaction loop.
At each step, the agent observes the current state, including the user query, available tools and their costs, previously obtained data, and feedback from earlier actions, then decides which atomic or composite tool to invoke next.
The environment responds with an updated state and cost-tagged feedback.
The interaction terminates once the agent produces the final output or reaches the predefined task sequence length.

\paragraph{Dynamic Blocking.} 
\cb{} includes an optional dynamic blocking module that introduces behavior-adaptive runtime perturbations to evaluate the robustness of cost-aware agents. The agent may encounter four types of disruptions, while preserving goal reachability (see Appendix~\ref{app:block-but-solvable}):
\begin{itemize}[topsep=2pt, partopsep=3pt, leftmargin=*, itemsep=-3pt]
\item \textbf{Explicit Blocks:}
disruptions with direct notification (e.g., tool/user message).
(1) \textit{Ban Tool}: behavior-dependent failures where specific tools become unavailable due to the agent’s prior actions.
(2) \textit{Preference Change}: modifications to user preferences that require the agent to replan its strategy mid-execution.
\item \textbf{Implicit Blocks:}
disruptions without explicit prompt.
(1) \textit{Cost Change}: global adjustments to tool costs that alter the current cost landscape.
(2) \textit{Remove Tools}: removes composite tools consisting of a certain number of component tools, thereby reducing the available action paths.
\end{itemize}
The mechanism ensures that:
\textit{(1) Dynamic and Uniform Triggering}: Blocking events are triggered dynamically based on estimated cost-optimal path lengths, yielding approximately uniform trigger positions and ensuring disruptions occur at meaningful stages (see Appendix~\ref{app:blocking-details}).
\textit{(2) Evaluation Fairness}: The mechanism ensures that any agent following the ground-truth optimal path encounters the same blocking events as others, guaranteeing fair comparison across models.
\textit{(3) Independence Across Data Instances}: For each run, environments are instantiated independently across data points, preventing correlations among blocking events. 
\textit{(4) Randomized Yet Reproducible Execution}: Even under dynamic conditions, experiments remain \textit{randomized yet reproducible}, preserving diversity and fairness in evaluation (see Appendix~\ref{app:randomized-yet-reproducible-environment}). 

\subsection{User Query Construction}
\label{sec:query-construction.}

For each of the six travel-planning tasks, we construct user queries through a multi-stage pipeline.
Each task defines four preference dimensions with ten candidate values, where four are used for test query construction and six for the training set, ensuring clear differentiation and non-overlapping splits, resulting in \(4^4 + 6^4 = 1{,}552\) unique user preference combinations per task.
All combinations are verified by \textit{GPT-4o} and partially checked manually for clarity (see Appendix~\ref{app:data-annotation}).
Natural language queries are generated by \textit{GPT-4o} to capture realistic vagueness while preserving clear distinctions among user preferences within each dimension.
An example query is shown in Figure~\ref{fig:example-user-query}, and dataset statistics are summarized in Tables~\ref{tab:benchmark-statistics_query} and~\ref{tab:benchmark-statistics_tool}.

\begin{table*}[t]
\begin{center}
\tabcolsep=0.015\linewidth
\resizebox{0.9\linewidth}{!}{
\begin{tabular}{lcccccc}
\toprule
\multicolumn{1}{c}{\multirow{2}{*}{\textbf{Model Name}}} & \multicolumn{1}{c}{\textbf{Cost}} & \multicolumn{3}{c}{\textbf{Path}} & \multicolumn{1}{c}{\textbf{Task Completion}} & \multicolumn{1}{c}{\textbf{Tool Call}} \\
\cmidrule(lr){2-2} \cmidrule(lr){3-5} \cmidrule(lr){6-6} \cmidrule(lr){7-7}

 & Cost Gap $\downarrow$ & AED $\downarrow$ & ANED (\%) $\downarrow$ & EMR (\%) $\uparrow$ & TCR (\%) $\uparrow$ & ITUR (\%) $\downarrow$ \\
\midrule
\textit{Greedy Policy} & 0.269 (0.269) & 2.202 & 74.74 & 10.76 & - & - \\
\textit{Qwen3-8B} & 0.477 (0.239) & 2.024 & 71.05 & 17.32 & 82.94 & \phantom{0}0.72 \\ 
\textit{Qwen3-14B} & 0.670 (0.130) & 1.976 & 51.26 & 33.45 & 88.71 & \phantom{0}0.91 \\ 
\textit{Qwen3-32B} & \underline{0.108} (0.108) & 1.748 & 52.51 & 33.60 & 90.55 & \phantom{0}1.73 \\
\textit{Llama-3.1-8B-Instruct} & 0.303 (0.303) & 2.215 & 80.21 & \phantom{0}9.71 & 63.25 & 32.50 \\
\textit{GLM-4.5} & 0.165 (0.066) & 1.076 & 34.79 & 54.33 & 90.55 & \phantom{0}4.22 \\ 
\textit{Deepseek-V3.1} & 2.221 (0.200) & 2.320 & 65.28 & 17.85 & 87.93  & 21.26 \\ 
\textit{Gemini-2.5-Pro} & 0.312 (\underline{0.015}) & \underline{0.367} & \underline{10.98} & \underline{83.73} & \textbf{94.49} & \textbf{\phantom{0}0.00} \\ 
\textit{Claude-Sonnet-4} & 0.249 (0.078) & 0.982 & 31.28 & 59.84 & \underline{93.58} & \phantom{0}5.04 \\ 
\textit{GPT-4o} & 1.179 (0.228) & 2.454 & 68.12 & 13.65 & 90.29 & 19.96 \\ 
\textit{GPT-5} & \textbf{0.060 (0.002)} & \textbf{0.103} & \textbf{\phantom{0}3.31} & \textbf{95.52} & 92.11 & \phantom{0}\underline{0.01} \\ 
\bottomrule
\end{tabular}
}
\end{center}
\vspace{-0.1in}
\caption{\CB{} main evaluation results for task sequences of length 5 in the \textit{static} setting (see results for task sequence of length 8 in Table~\ref{tab:more_unblocked_results}). 
Scores in \textbf{bold} indicate the best performance among all models, and \underline{underlined} scores denote the second-best performance.
The metrics in (\%) are in percentage forms. Reported values under \textbf{Cost Gap} indicate the average cost gap, with numbers in parentheses computed after excluding samples with redundant tool calls (see Appendix~\ref{app:error-analysis} for details).
The greedy baseline selects the atomic tool with the lowest average cost at each step (See Appendix~\ref{app:details-gt-greedy}) and the average tool call number is provided in Table~\ref{tab:tool_use_number_statistics}.}
\label{tab:main_unblocked_results}
\vspace{-0.15in}
\end{table*}

\begin{figure*}[ht]
    \centering
    \includegraphics[width=\linewidth]{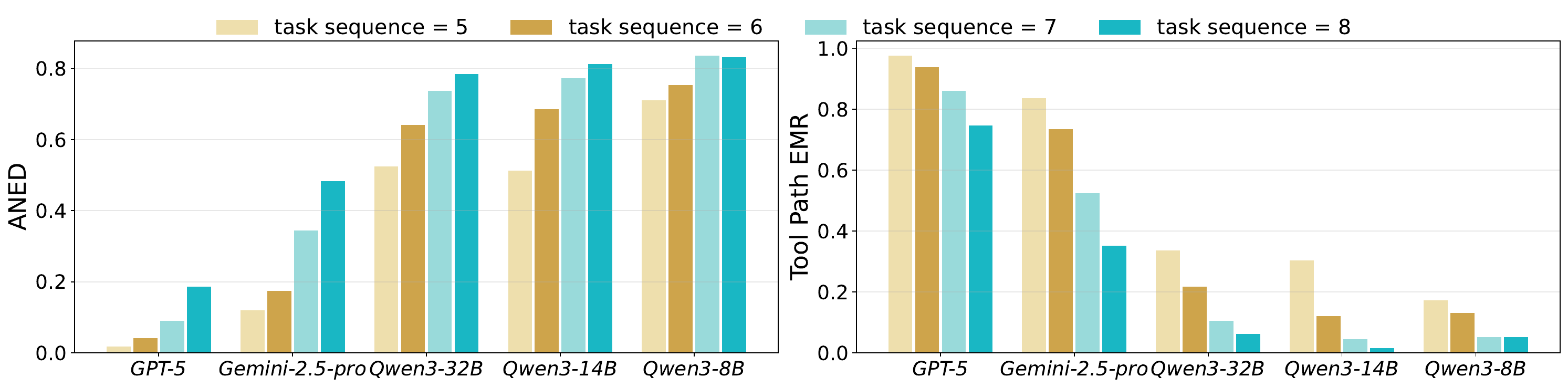}
    \vspace{-0.3in}
    \caption{Models’ average normalized edit distance and exact match ratio for task sequences of length five to eight. Performance drops significantly as complexity increases. At length eight, even the strongest model, \textit{GPT-5} achieves less than 75\% exact match (see Appendix~\ref{app:task-sequence-8-analysis} for more results).}
    \label{fig:unblocked_scaling_combined}
    \vspace{-0.1in}
\end{figure*}


\section{Experiment}

\subsection{Settings}
\label{sec:experiment-settings}

\paragraph{Models.}
We evaluate both proprietary and open-source models to ensure a balanced and comprehensive assessment of current LLM capabilities.
Proprietary models include 
\textit{GPT}~\cite{gpt-4o}, 
\textit{Claude}~\cite{claude}, 
\textit{Deepseek}~\cite{deepseek-v3.1}, 
and \textit{Gemini}~\cite{gemini}; 
while open-source models include 
\textit{GLM}~\cite{glm-4.5}, 
\textit{Qwen3}~\cite{qwen3}, 
and \textit{Llama3}~\cite{Llama3.1}. 
All models use a temperature of 0.0 for deterministic decoding and a maximum token length of 16,384 to prevent output truncation.

\paragraph{Metrics.}\label{sec:metrics}

We evaluate performance using metrics spanning four categories: cost, path, task completion, and tool use. Most metrics measure the deviation between an agent’s trajectory and the ground-truth trajectory, with a greedy policy as the baseline. Full metric definitions and illustrative examples are provided in Appendix~\ref{app:metrics-implementation} and Appendix~\ref{app:metrics-example}.

\noindent \textbf{(1) Cost Gap:} This metric directly measures the gap between the ground-truth and agent accumulated explicit tool costs.

\noindent \textbf{(2) Average Edit Distance (AED):} This metric measures the edit distance between the ground-truth and the agent’s trajectories to assess the quality of the agent’s path.

\noindent \textbf{(3) Average Normalized Edit Distance (ANED) (\%):} This metric normalizes the ED to [0, 1] for fair comparison.

\noindent \textbf{(4) Exact Match Ratio (EMR) (\%):} The ratio of data points that agent implements an identical tool call trajectory as ground truth.

\noindent \textbf{(5) Task Completion Ratio (TCR) (\%):} The proportion of cases where the agent returns the unique correct answer, regardless of tool path.

\noindent \textbf{(6) Invalid Tool-Use Ratio (ITUR) (\%)} This metric evaluates the proportion of invalid tool use. Invalid behavior includes using incorrect tool names, incorrect parameter names, and similar errors.

\paragraph{Prompts and Environment Settings.}

Detailed prompt and environment hyperparameters are in Figure~\ref{fig:runtime-prompt} and Table~\ref{tab:environment-hyperparas}. To prevent trivial one-step solutions, tools that complete the entire procedure in a single call are excluded.

\subsection{Results}


\paragraph{Most models perform poorly in cost-optimal planning.} 

As shown in Table~\ref{tab:main_unblocked_results}, while most models complete the task well, only \textit{GPT-5} and \textit{Gemini-2.5-Pro} show strong exact-match performance at a task length of five, whereas all other models fall below 60\%. 
Performance further declines for longer task sequences (see Section~\ref{sec:analysis-for-static-setting}), where even \textit{GPT-5} drops to below \textbf{75\%} EMR.
This highlights a substantial gap in executing complete cost-aware plans, as even the strongest model, \textit{GPT-5}, falls short of reliably producing optimal trajectories. 
Second, compared to the greedy baseline, some LLMs fail to show convincing improvements. For example, \textit{Qwen3-8B}, \textit{GPT-4o}, and \textit{Deepseek-V3.1} yield nearly identical ANED scores to greedy policy, with \textit{GPT-4o} even exhibiting a higher AED, indicating that its generated plans deviate further from optimal solution than greedy baseline.
More concerningly, \textit{Llama-3.1-8B-Instruct} underperforms the greedy baseline across all path-related metrics, suggesting fundamental weaknesses in cost-optimal planning. Robustness checks for our evaluation are detailed in Appendix~\ref{app:effect-of-random-seeds}.

\begin{figure*}[ht]
\centering
    \begin{minipage}{0.5\textwidth}
        \centering
        \includegraphics[width=\textwidth]{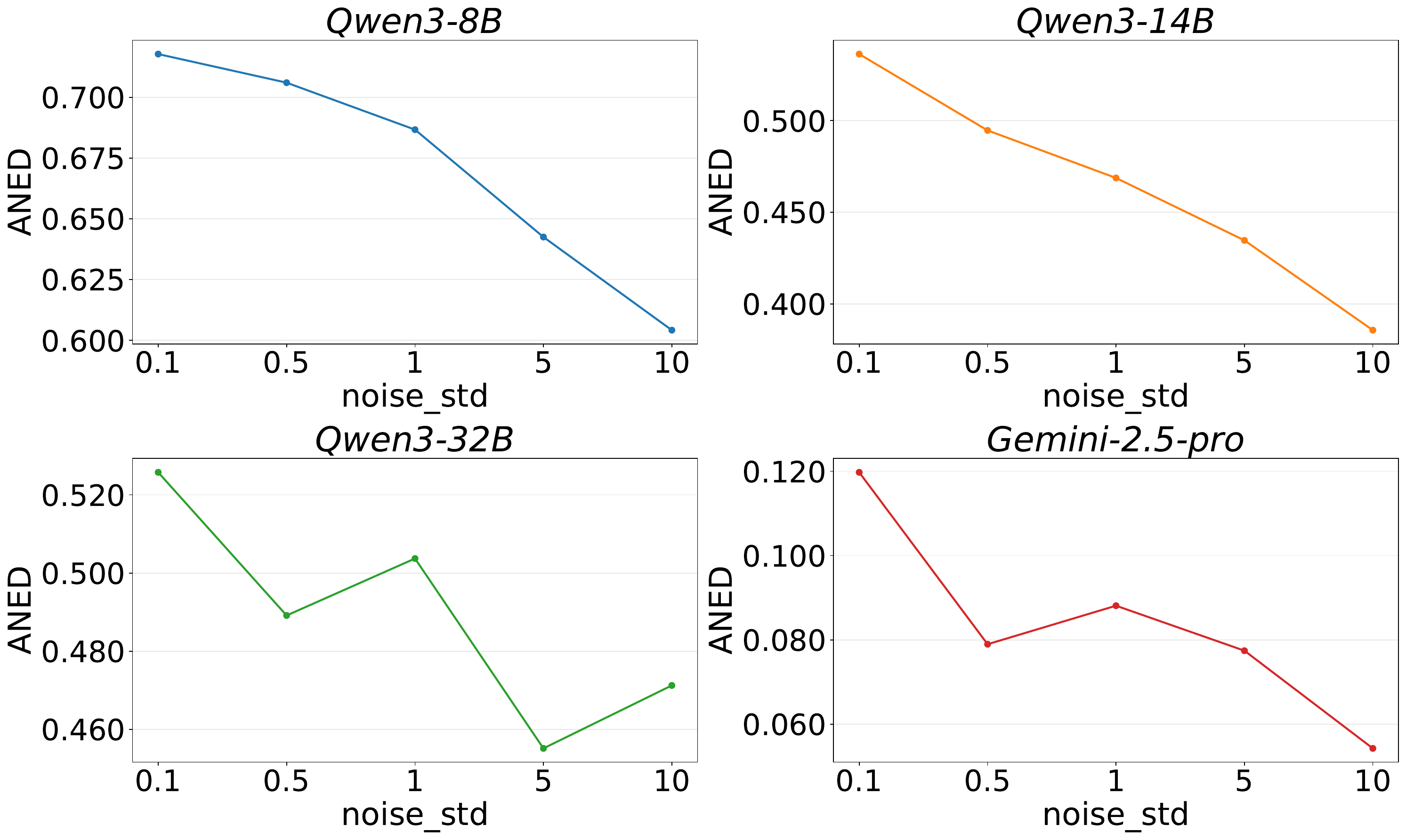}
    \end{minipage}%
    \hfill
    \begin{minipage}{0.5\textwidth}
        \centering
        \includegraphics[width=\textwidth]{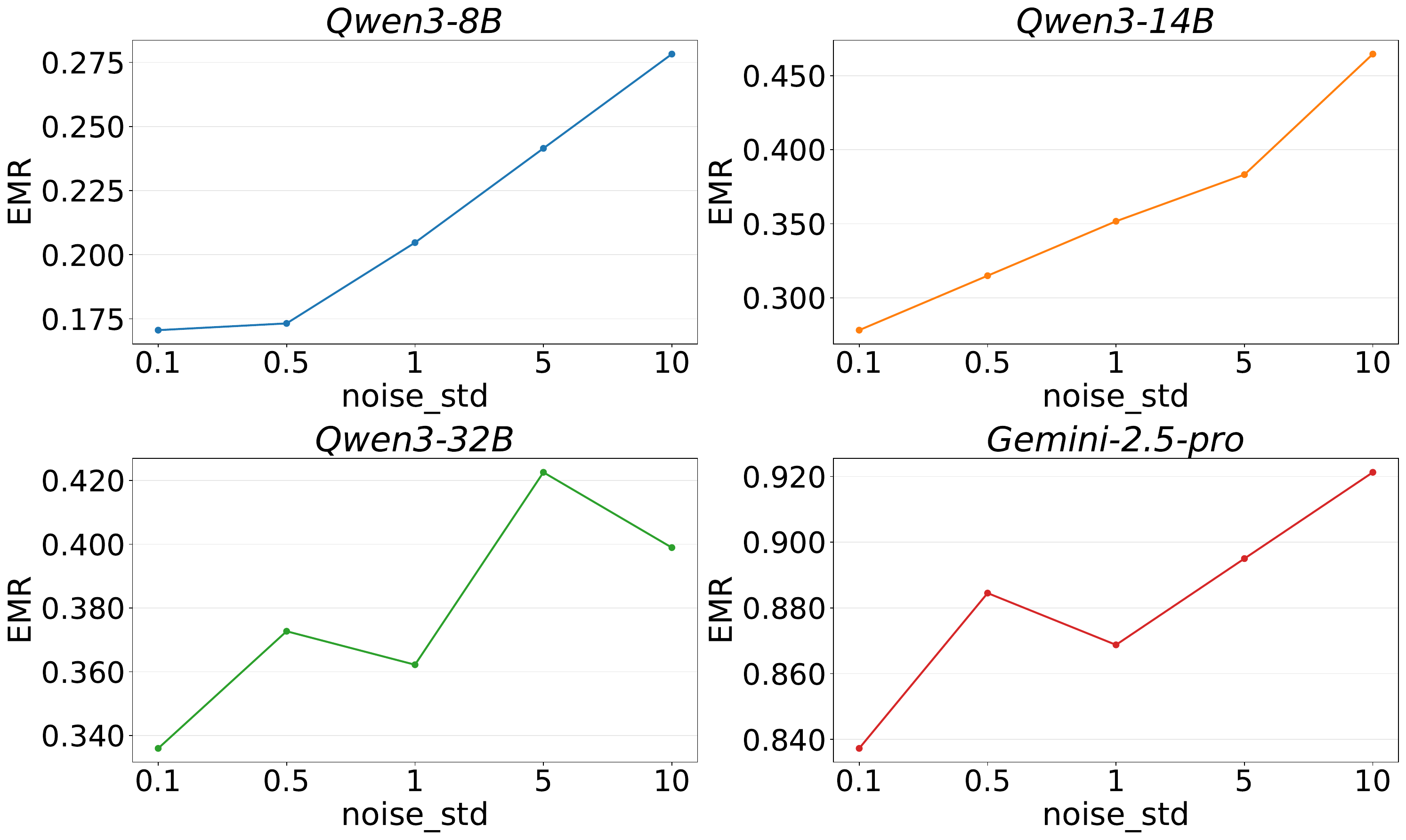}
    \end{minipage}
    \vspace{-0.1in}
    \caption{LLMs’ performance on \cb{} under different standard deviations of composite tool cost noise. Models tend to perform better with higher noise levels, indicating high sensitivity to tool cost variations.}
    \label{fig:noise-ablation}
    \vspace{-0.1in}
\end{figure*}

\begin{figure}[ht]
    \centering
    \includegraphics[width=\linewidth]{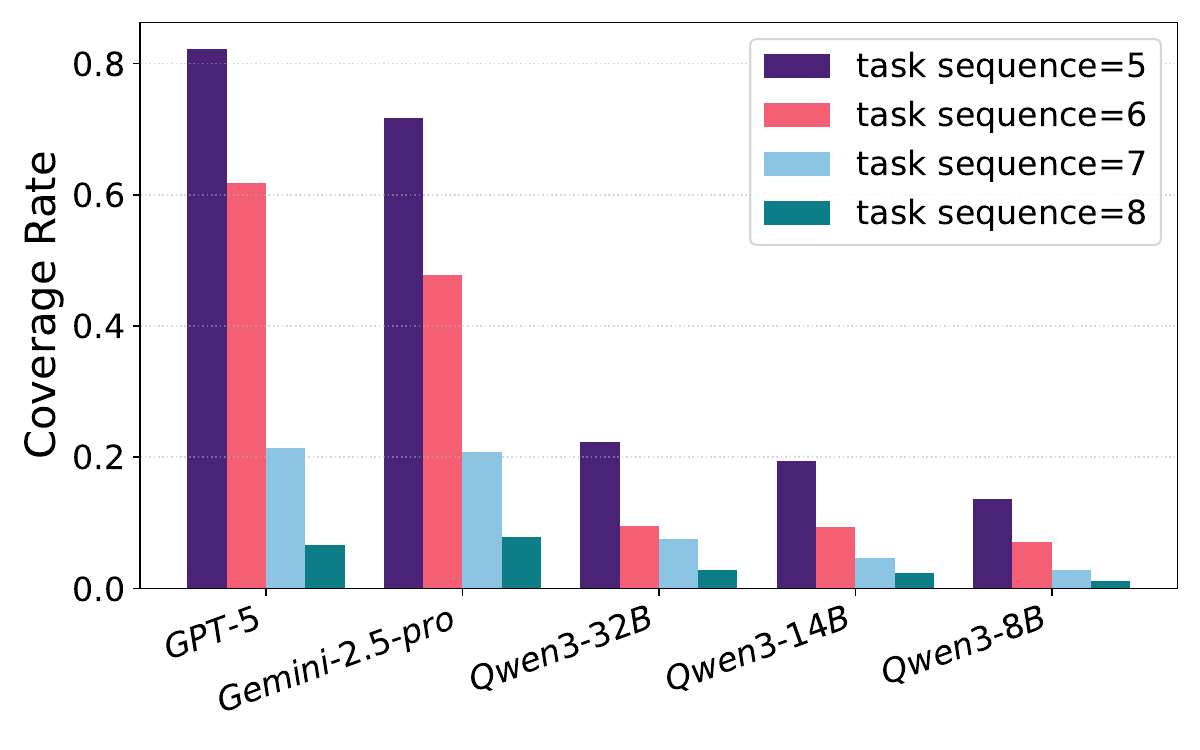}
     \vspace{-0.33in}
    \caption{Coverage rates of different LLMs across task sequences of length 5 to 8. Model performance in the static setting shows a strong positive correlation with coverage rate.} 
    \label{fig:coverage-analysis}
    \vspace{-0.15in}
\end{figure}

\paragraph{Model suffers from redundant tool calls and tool call failures.} In Table~\ref{tab:main_unblocked_results}, we report two variants of the cost gap metric: values outside the parentheses are computed over all samples, while those inside exclude two abnormal cases—\textit{repeated tool calls} and \textit{extra tool calls}. After removing these anomalies, the cost metrics align with the path-related metrics in reflecting model performance. Beyond such redundant calls, we also observe failure cases including \textit{invalid parameter inputs} and attempts to invoke \textit{inaccessible tools}, the latter being the most severe. These errors reveal models’ limited progress awareness, which in turn hinders their ability to plan in a cost-optimal manner.
Further discussion of the failure modes can be found in Appendix~\ref{app:error-analysis}.

\paragraph{Comparison of Open-Source and Proprietary Models.} Beyond the greedy baseline, we compare open-source and proprietary models. Interestingly, some proprietary models show no clear advantage~\cite{ACE-Bench}: \textit{Qwen3-32B} consistently surpasses \textit{GPT-4o} and \textit{DeepSeek-V3.1} on all path-related metrics, and \textit{GLM-4.5} achieves performance comparable to \textit{Claude-Sonnet-4}. While partly due to differing invalid tool-use rates, our metrics decouple tool use failure from cost or path measurements (see Appendix~\ref{app:metrics-boundary-conditions}), indicating that some proprietary models’ shortcomings in cost-sensitive planning are substantive and reflect genuine limitations in their planning capabilities.

\paragraph{Limited gains from parameter scaling in Qwen models.}
Scaling up model size leads to performance improvements, but the gains are limited. Within the Qwen family, moving from \textit{Qwen3-8B} to \textit{Qwen3-14B} yields a clear boost, whereas the improvement from \textit{Qwen3-14B} to \textit{Qwen3-32B} is marginal. This indicates diminishing returns from parameter scaling alone, pointing to the need for algorithmic or architectural advances beyond sheer model size.

\section{Analysis}\label{sec:analysis}

\subsection{Static Setting}
\label{sec:analysis-for-static-setting}

\paragraph{Model performance declines as task complexity increases.}
As shown in Figure~\ref{fig:unblocked_scaling_combined}, performance consistently deteriorates as task sequences grow longer.
The best-performing model, \textit{GPT-5}, drops from near-perfect exact match ratio at length 5 to around 75\% at length 8, accompanied by a higher normalized edit distance, while weaker models such as \textit{Qwen3-8B} and \textit{Qwen3-14B} collapse entirely (EMR = 0).
This overall downward trend highlights the increasing difficulty current models face in sustaining coherent, cost-optimal reasoning as task complexity rises.

\paragraph{Coverage Rate in planning significantly influences performance.}\label{sec:coverage-rate}
In the unblocking setting, the model’s task is to enumerate all possible paths and select the optimal one. 
We define coverage as the proportion of paths explicitly planned, and hypothesize that higher coverage leads to better performance. 
We employ \textit{GPT-4o-mini}~\cite{GPT-4o-mini} to extract distinct paths from each model’s output (prompt in Figure~\ref{fig:extract-converage-prompt}, human check in Appendix~\ref{app:coverage-rate-human-check}).
As shown in Table~\ref{tab:main_unblocked_results}, performance follows \textit{GPT-5} $>$ \textit{Gemini-2.5-pro} $>$ open-source \textit{Qwen} models, which aligns with the coverage trend in Figure~\ref{fig:coverage-analysis}. 
Performance also declines as task sequences grow longer (Figure~\ref{fig:unblocked_scaling_combined}), explained by the corresponding drop in coverage. 
Importantly, even when given abstract examples that enumerate all paths (Figure~\ref{fig:runtime-prompt-two}), models often fail to generalize these planning strategies to real scenarios.

\paragraph{Models exhibit pronounced cost sensitivity.}

To examine models’ sensitivity to cost variations, we vary the noise standard deviation when constructing composite tools, which controls how much a composite tool’s cost deviates from the sum of its component tools.
Larger noise introduces greater diversity in cost structures and reshapes the cost landscape.
As shown in Figure~\ref{fig:noise-ablation}, model performance, measured by both ANED and EMR, consistently improves as the noise increases.
This result suggests that models are highly responsive to cost signals, as greater cost differences between action paths make it easier for them to identify the optimal path without explicitly computing the total cost of each option.
\vspace{-0.05in}

\subsection{Dynamic Setting}

\paragraph{Models remain highly vulnerable to dynamic blocking.}

\begin{figure*}
    \centering
    \includegraphics[width=\linewidth]{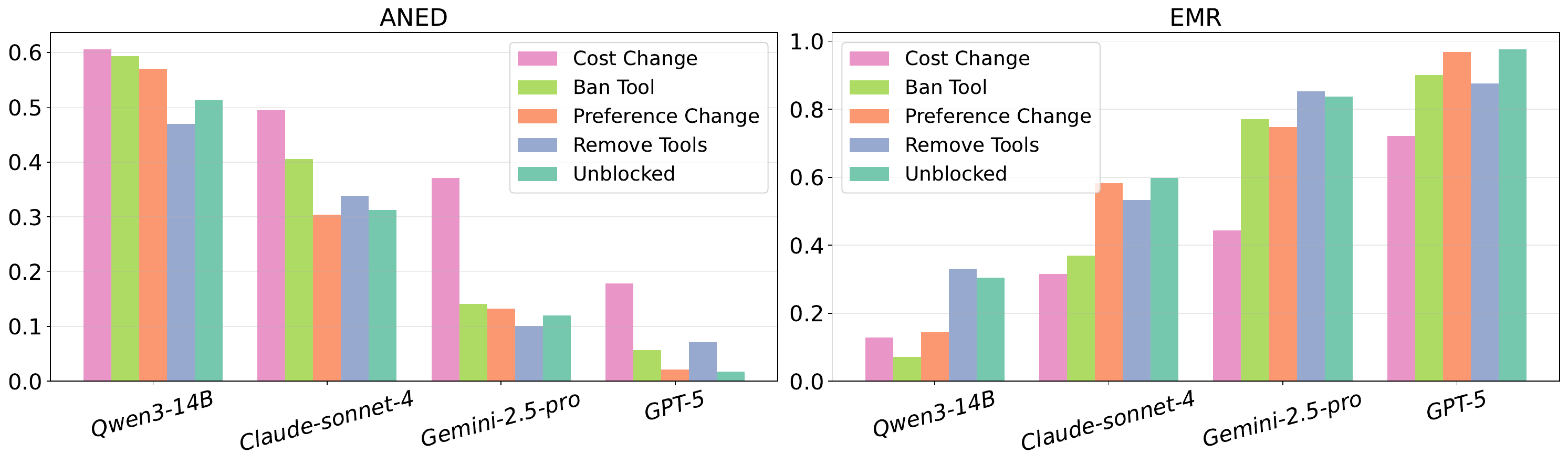}
    \vspace{-0.2in}
    \caption{LLMs’ performance in \cb{}’s dynamic blocking setting. All models show consistent EMR drops under ban tool, cost change, and preference change conditions, indicating weak replanning and adaptation abilities in dynamic environments.}
    \label{fig:block_comparison}
    \vspace{-0.1in}
\end{figure*}

\begin{figure*}[ht]
\centering
    \begin{minipage}{0.5\textwidth}
        \centering
        \includegraphics[width=\textwidth]{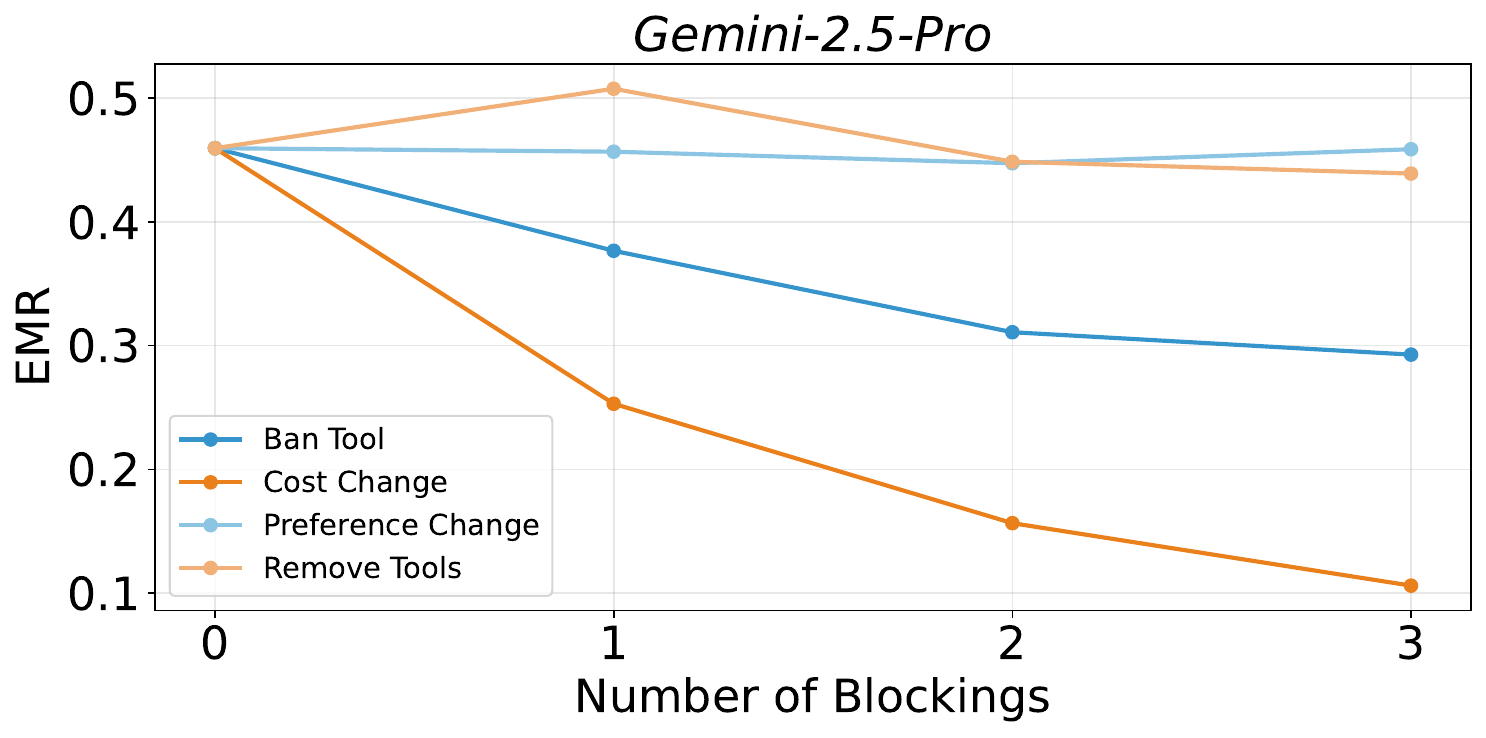}
    \end{minipage}%
    \hfill
    \begin{minipage}{0.5\textwidth}
        \centering
        \includegraphics[width=\textwidth]{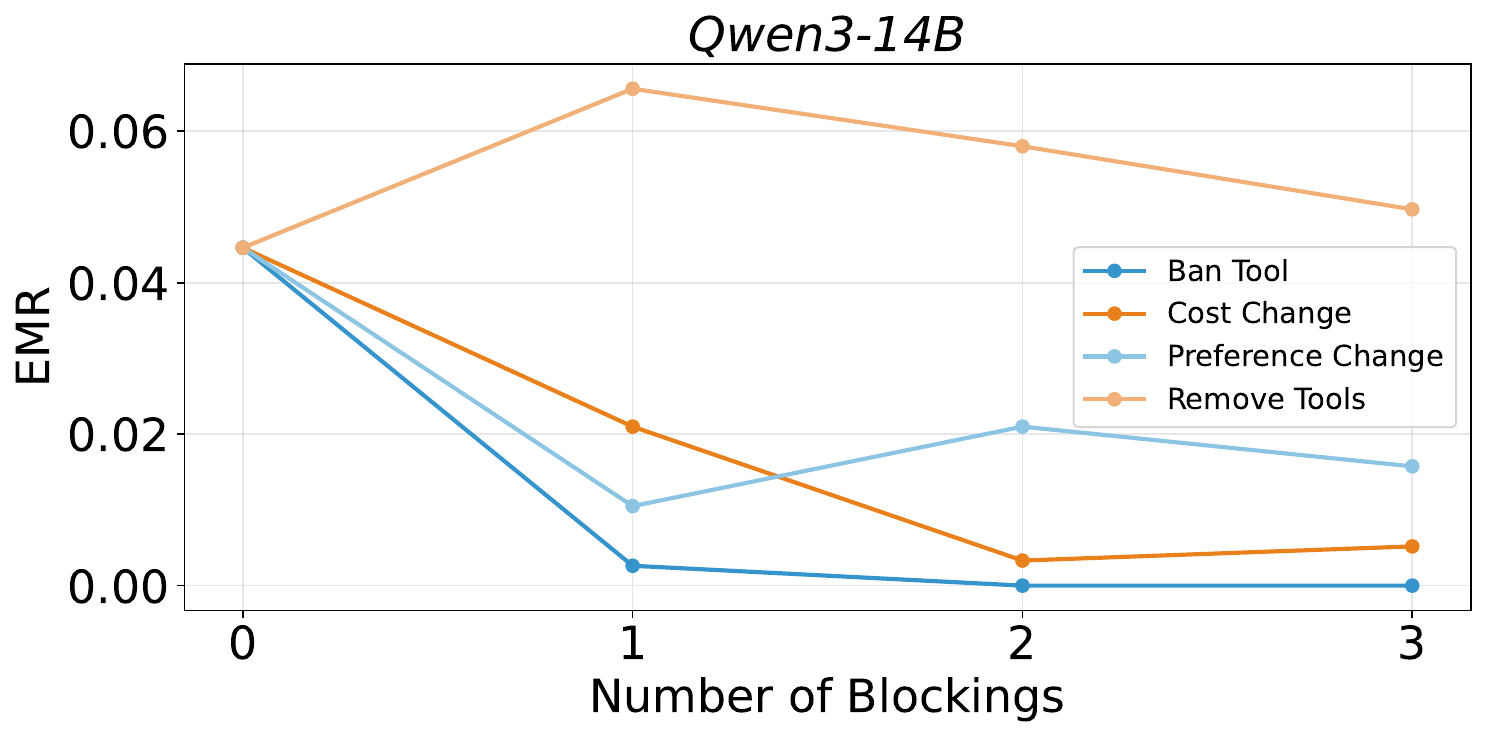}
    \end{minipage}
    \vspace{-0.1in}
    \caption{Performance of \textit{Gemini-2.5-Pro} and \textit{Qwen3-14B} under increasing numbers of blocking events (task sequence length = 7). 
    Each curve represents a different blocking type. Both models degrade with more blockings, especially under frequent cost changes or tool bans, with \textit{Qwen3-14B} showing near-total failure.}
    \label{fig:block-scale}
    \vspace{-0.1in}
\end{figure*}

From Figure~\ref{fig:block_comparison}, we observe that dynamic blocking substantially affects both ANED and EMR across all models. 
Among the blocking types, \textit{cost change} and \textit{ban tool} consistently induce the most significant degradation. 
For \textit{cost change}, the EMR of \textit{GPT-5}, which originally achieved nearly perfect EMR, drops by around 20 percentage points, while \textit{Gemini-2.5-Pro} suffers an even larger decline of almost 40 points. 
To further examine model robustness in the most challenging cases, we evaluate \textit{GPT-5} under the hardest \textit{cost-change} setting, where its Exact Match falls to merely \textbf{34.91\%}. Under the \textit{ban tool} condition, both \textit{GPT-5} and \textit{Gemini-2.5-Pro} still experience noticeable drops (approximately 5--10\%), yet they exhibit stronger resilience compared to the other two models. In addition, \textit{preference change} also causes a stable decrease in Exact Match across all evaluated models. 
In contrast, the effect of \textit{remove tools} is less consistent, which we attribute to the reduced search space allowing models to achieve higher coverage and thus more stable performance (see Section~\ref{sec:coverage-rate} for the coverage–performance relationship).

To better understand performance degradation across blocking types, we estimate the relative adaptation difficulty by measuring how much ground-truth trajectories change under each condition, using the average normalized edit distance (ANED) between unblocked and blocked settings.
Higher ANED indicates greater deviation in the optimal path and thus higher adaptation difficulty.
Results show that \textit{cost change} (mean$_{\text{GT-NED}}$ = 0.378) and \textit{ban tool} (0.538) cause larger deviations than \textit{preference change} (0.295) and \textit{remove tools} (0.214), aligning with the stronger performance drops in Figure~\ref{fig:block_comparison}.
Notably, although \textit{ban tool} produces greater trajectory deviation, \textit{cost change} yields the steepest performance decline, suggesting that implicit blocking (e.g., cost perturbation) poses extra challenges, as models must first detect the change before replanning—revealing their limited adaptability.

\begin{table*}[t]
\begin{center}
\tabcolsep=0.015\linewidth
\resizebox{0.8\linewidth}{!}{
\begin{tabular}{lcccccc}
\toprule
\multicolumn{1}{c}{\multirow{2}{*}{\textbf{Model Name}}} & \multicolumn{1}{c}{\textbf{Cost}} & \multicolumn{3}{c}{\textbf{Path}} & \multicolumn{1}{c}{\textbf{Task Completion}} & \multicolumn{1}{c}{\textbf{Tool Call}} \\
\cmidrule(lr){2-2} \cmidrule(lr){3-5} \cmidrule(lr){6-6} \cmidrule(lr){7-7}

 & Cost Gap $\downarrow$ & AED $\downarrow$ & ANED (\%) $\downarrow$ & EMR (\%) $\uparrow$ & TCR (\%) $\uparrow$ & ITUR (\%) $\downarrow$ \\
\midrule
\textit{Greedy Policy} & 0.524 (0.524) & 3.194 & 84.82 & \phantom{0}3.41 & - & - \\
\textit{Qwen3-8B} & 1.027 (0.496) & 3.199 & 83.19 & \phantom{0}5.25 & 85.30 & 2.92 \\ 
\textit{Qwen3-14B} & 0.503 (0.475) & 5.564 & 81.28 & \phantom{0}1.57 & 88.71 & \underline{0.48} \\ 
\textit{Qwen3-32B} & 0.525 (0.393) & 4.068 & 78.49 & \phantom{0}6.30 & 88.83 & 0.71 \\ 
\textit{Gemini-2.5-Pro} & \textbf{0.127} (\underline{0.116}) & \underline{2.072} & \underline{48.25} & \underline{35.15} & \underline{91.13} & \textbf{0.01} \\ 
\textit{GPT-5} & \underline{0.311} (\textbf{0.029}) & \textbf{0.782} & \textbf{18.70} & \textbf{74.79} & \textbf{91.60} & 1.80 \\ 
\bottomrule
\end{tabular}
}
\end{center}
\vspace{-0.1in}
\caption{\CB{} main evaluation results of selected models for task sequences of length 8 in the \textit{static} setting. 
Similar to Table~\ref{tab:main_unblocked_results},  scores in \textbf{bold} and \underline{underlined} indicate the best and second-best performance, respectively.
The metrics in (\%) are in percentage forms. 
Reported values under \textbf{Cost Gap} indicate the average cost gap, with numbers in parentheses computed after excluding samples with redundant tool calls (see Appendix~\ref{app:error-analysis} for details).
The greedy baseline selects the atomic tool with the lowest average cost at each step (See Appendix~\ref{app:details-gt-greedy}). 
}
\label{tab:more_unblocked_results}
\vspace{-0.15in}
\end{table*}

\paragraph{Multiple blockings severely undermine model robustness.}

We further analyze the results under multiple blocking events (Figure~\ref{fig:block-scale}), focusing on how repeated disruptions impact model robustness and cost-optimal planning.
We find that increasing the frequency of both \textit{ban tool} and \textit{cost change} blocks leads to a substantial drop in EMR across all models. 
For instance, the EMR of \textit{Gemini-2.5-Pro}, which initially achieved around 45\% accuracy, drops to nearly 0.1 after three consecutive \textit{cost-change} events, while \textit{Qwen3-14B} completely fails (EMR = 0) after only two such events. 
This indicates that even state-of-the-art commercial and open-source models struggle to maintain cost-awareness under complex dynamic environments, highlighting their limited robustness and adaptability to repeated disruptions.
\section{Discussions}
\label{sec:discussion}

\paragraph{On the Significance of \CB{}.}

Evaluating cost-aware planning in LLM agents has broad applicability across real-world scenarios. 
First, virtually all tool-use contexts inherently involve costs, as invoking tools in practical applications typically incurs resources such as API fees for search engines~\cite{JinaAI} or Model Context Protocol (MCP) services~\cite{openrouter}. 
While tool integration represents a major advancement distinguishing agents from standalone LLMs, with paradigms like Tool-Integrated Reasoning (TIR) gaining widespread adoption~\cite{toolrl,Tool-R0}. 
Thus, equipping agents with cost awareness during tool use is essential for promoting efficient, economically rational behavior in resource-constrained environments.
Second, cost awareness has been empirically shown to enhance overall agent performance. 
As demonstrated by~\citet{budget-aware-tool-use}, budget awareness serves as a key driver for improving agent capabilities, particularly in scaling up agents for long-horizon tasks. 
In such contexts, cost awareness emerges as a powerful aid for test-time scaling, enabling agents to optimize trajectories, reduce redundant calls, and adapt dynamically to achieve better outcomes under prolonged interactions.

\paragraph{On the Generalization of \CB{}.}

Although \cb{} is situated in the travel-planning domain, its underlying linear tool structure, as demonstrated in Figure~\ref{fig:tool-flow-example}, mirrors patterns prevalent across a wide array of tasks~\cite{wen2020hierarchical}.
Most real-world applications inherently involve consecutive steps~\cite{schneider2006hierarchical}, where sequential dependencies create opportunities for cost-aware optimization challenges similar to those in \cb{}. 
Furthermore, \cb{}'s emphasis on comparing equivalent paths highlights a common dilemma in real-world MCP servers where agents must select the most suitable tool from functionally similar options to minimize costs while achieving the same outcome~\cite{blankenstein2026biasbustersuncoveringmitigatingtool}. 
This design not only tests agents' economic reasoning but also promotes generalizable strategies for efficient tool selection in diverse, dynamic environments.
\section{Conclusion}
We present CostBench, a scalable, cost-focused benchmark for evaluating LLM agents’ ability to plan and adapt cost-optimally in dynamic environments. Centered on travel planning, it includes atomic and composite tools with randomized costs and four types of dynamic blocking events that force adaptive replanning. Evaluating ten leading LLMs, we find substantial weaknesses: even the best-performing \textit{GPT-5} achieves under 75\% exact match on the hardest static tasks, with performance dropping to around 35\% under cost-change conditions and showing similar vulnerability when tools are banned.
These results reveal limitations in current models’ cost-aware reasoning, path enumeration, and adaptability. Beyond diagnosing these weaknesses, CostBench provides a principled framework to drive the next generation of LLM agents toward economically rational, resource-efficient, and resilient decision-making in complex, real-world scenarios.

\section*{Limitations}
While \cb{} provides a principled and scalable framework for evaluating cost-aware planning, several limitations remain.
First, our current tasks are restricted to the travel-planning domain, which, though diverse, may not fully capture the breadth of real-world cost trade-offs.
Nevertheless, \cb{} can be easily extended to other domains by adding configuration files.
Second, our simulation abstracts away true API latency, resource consumption, and stochastic failures, which may further influence agent behavior in real-world deployment.
Third, the blocking events we design, while representative, are still manually parameterized; future work could explore learning-based or user-driven dynamics to better approximate natural environmental shifts.

\section*{Ethics Statement}

\paragraph{Offensive Content Elimination.} 
Our benchmark focuses on the travel-planning domain, and its curation does not rely on content generated by LLMs. Instead, all data were carefully sampled and validated to ensure the dataset is free of offensive material. Consequently, we are confident that it poses no risk of negative societal impact.

\paragraph{Licenses.} 
Our code will be released under the MIT license to allow unrestricted research use. The \cb{} will be distributed under a Creative Commons (CC) license, providing free access for the academic community. Our use of existing models and tools is strictly consistent with their original licenses and intended research purposes. We take full responsibility for any potential rights violations or licensing issues, and all resources comply with their respective terms of use while supporting research purposes.

\paragraph{Models.} 
All open-source models were hosted and executed locally using the vLLM library~\cite{VLLM}, while all closed-source models were accessed through their respective official APIs. 
For reproducibility, the experimental settings are detailed in Section~\ref{sec:experiment-settings}.

\paragraph{Data Annotations.}

All data annotation was performed by the paper's co-authors, who are qualified researchers with relevant expertise, ensuring that the process was conducted responsibly and in accordance with ethical standards.


\bibliography{custom}

\clearpage
\appendix

\section{Comparison Traits Details}
\label{app:Comparison-of-related-work}

We distill eight core traits that capture the essential characteristics of recent benchmarks designed for cost-optimal evaluation. These traits concern both the agent’s ability to conduct realistic, multi-step, and efficiency-oriented interactions, as well as the infrastructure’s capacity to enable scalable and extensible evaluation. 
\begin{itemize}
    \item \textbf{Multi-turn Interaction}: The benchmark supports and requires multi-turn dialogue or decision-making. This trait is essential, as recent agentic paradigms increasingly emphasize agent performance in multi-turn interactions. From web search~\cite{webwatcher,webdancer,liu-etal-2024-gprooft} to GUI-based scenarios~\cite{GUI-agent-multiturn,UI-TARS-2}, agentic performance can only be thoroughly evaluated in dynamic, multi-turn environments.
    \item \textbf{Cost-optimal Planning}: Agent performance is explicitly assessed by cost-optimality. We distinguish \emph{budget awareness}, or maximizing performance under a fixed budget, from \emph{cost-optimality}, which requires reaching the goal with the minimum possible cost. 
    \item \textbf{Flexible Cost}: The environment allows user-defined costs or provides sufficiently diverse cost combinations. As discussed in Section~\ref{sec:intro}, a more flexible and diverse cost assignment is crucial for accurately evaluating an agent’s genuine cost awareness, such as its sensitivity to varying costs.
    \item \textbf{Tool Use}: The benchmark incorporates or necessitates the use of external tools. This trait is particularly important, as tool use has emerged as a defining capability of modern LLM agents—central to their ability to interact with and act upon the external world~\cite{toolsandbox, MINT, ToolBench,UltraTool,qian2026advancing,qian2026creativitybench}. Evaluating tool use is thus indispensable for understanding an agent’s true reasoning and decision-making competence.
    \item \textbf{Dynamic State}: The environment introduces runtime-controllable obstacles that may block the agent and enforce replanning. This design is important because traditional benchmarks often suffer from data leakage risks~\cite{RFMBench,RL-data-contamination,survey-on-data-contamination,ConTAM}. In contrast, a dynamic environment can mitigate such risks by introducing runtime factors during evaluation, ensuring a fairer and more reliable assessment~\citep{liu2026planbench,liu2026adaplanbench}.
    \item \textbf{Adjustable Difficulty}: The difficulty level can be systematically adjusted, for example by scaling task parameters. A naturally adjustable difficulty hierarchy allows systematic evaluation of model adaptability under varying levels of challenge~\cite{UserBench,ComparisonQA,Math-perturb}, facilitating more fine-grained analysis.

\begin{figure}[t]
\centering
\begin{tikzpicture}[
    node distance=1.5cm,
    state/.style={rectangle, rounded corners, minimum width=4cm, minimum height=0.7cm, text centered, draw=black, fill=blue!10, font=\small},
    tool/.style={text centered, font=\scriptsize, align=center},
    arrow/.style={->, >=stealth, thick},
    composite/.style={->, >=stealth, thick, dashed, color=teal}
]

\node[state] (init) {Initial State};
\node[state, below of=init] (pref) {LocationPreference};
\node[state, below of=pref] (raw) {LocationCandidate\_Raw};
\node[state, below of=raw] (l1) {LocationCandidate\_L1};
\node[state, below of=l1] (final) {TravelLocation};

\draw[arrow] (init) -- node[tool, right, xshift=0.3cm] {Tool 1} (pref);
\draw[arrow] (pref) -- node[tool, right, xshift=0.3cm] {Tool 2} (raw);
\draw[arrow] (raw) -- node[tool, right, xshift=0.3cm] {Tool 3} (l1);
\draw[arrow] (l1) -- node[tool, right, xshift=0.3cm] {Tool 4} (final);

\draw[composite] (init.west) to[out=180, in=180] node[tool, left, xshift=1.0cm, yshift=-0.9cm] {Tool 5} (raw.west);
\draw[composite] (init.west) to[out=180, in=180] node[tool, left, xshift=1.3cm, yshift=-1.7cm] {Tool 6} (l1.west);
\draw[composite] (init.west) to[out=180, in=180] node[tool, left, xshift=1.75cm, yshift=-2.6cm] {Tool 7} (final.west);

\end{tikzpicture}

\vspace{0.2cm}
\noindent\footnotesize
\begin{flushleft}
\textbf{Atomic Tools:} \\
\textbf{Tool 1:} \texttt{Decide\_Location\_Preference}; \\
\textbf{Tool 2:} \texttt{Search\_Location\_Candidates}; \\
\textbf{Tool 3:} \texttt{Location\_Refinement\_Step1}; \\
\textbf{Tool 4:} \texttt{Select\_Final\_Location}

\vspace{0.1cm}
\textbf{Composite Tools:} \\
\textbf{Tool 5:} \texttt{Location\_Preference\_and\_Search} (= Tool 1+2); \\
\textbf{Tool 6:} \texttt{Location\_Full\_Planning\_to\_Step1} (= Tool 1+2+3); \\
\textbf{Tool 7:} \texttt{Location\_All\_pipeline} (= Tool 1+2+3+4) \\
\textbf{......}
\end{flushleft}

\caption{Example of sequential tool execution flow in \cb{} (task sequence = 4, requiring only one refinement step). Each atomic tool transforms one data type to another, ensuring explicit intermediate states and preventing shortcuts. In this figure, we illustrate only the composite tool starting from the initial state. In practice, a composite tool may commence from any state, encapsulating a sequence of atomic tool operations.}
\label{fig:tool-flow-example}
\end{figure}

\begin{table}[t]
    \centering
    \small
    \begin{tabular}{@{} l c @{}} 
        \toprule
        \textbf{Tools} & 
        \textbf{Value} \\
        \midrule
        Task Sequence & $N$ \\
        Total Atomic Tools & 6$N$ \\
        Total Composite Tools & 6${N*(N-1)/2}$ \\
        Total Tools & 6${N*(N+1)/2}$ \\
        
        \bottomrule
    \end{tabular}
    \vspace{-0.05in}

    \par
    \caption{Main statistics of tools used in \cb{}. The variable $N$ denotes an adjustable task sequence, and the total numbers of atomic and composite tools vary accordingly, as indicated (we use $N=5$ in most of our experiments).}
     \label{tab:benchmark-statistics_tool}
    \vspace{-0.05in}
\end{table}

\begin{table}[h!]
\centering
\small
\begin{tabular}{lc}
\toprule
\textbf{Parameter} & \textbf{Value} \\
\midrule
Task length & 5 \\
Maximum tool steps & 20 \\
Global seed ($S$) & 42 \\
Min atomic cost ($c_{\min}$) & 15 \\
Max atomic cost ($c_{\max}$) & 25 \\
Noise standard deviation ($\sigma$) & 0.1 \\
\bottomrule
\end{tabular}
\caption{Environment details for Table~\ref{tab:main_unblocked_results}. $c_{\min}$–$c_{\max}$: range for randomizing atomic tool costs (see Figure~\ref{fig:overview-pipeline}). noise std: standard deviation of Gaussian noise added to composite tool costs. The global seed $S$ is used to control the random seed for all data points in a given run, ensuring full reproducibility.}
\label{tab:environment-hyperparas}
\end{table}

    \item \textbf{Customized Details}: The environment offers fine-grained configurability of details. Enabling this feature makes the benchmark adaptable to other domains and facilitates broader applicability across diverse evaluation settings~\cite{toolsandbox, MINT}.
    \item \textbf{Scalability}: The benchmark can be automatically expanded in size without additional human annotation and easily generalized to other tasks. This property is crucial, as scalability and expandability are essential for large-scale evaluation~\cite{UserBench,tau-bench,liu2026naacl}.
\end{itemize}

\section{Experiment Details}

In this section, we provide a comprehensive overview of the experimental setup and evaluation methodology. 
We begin by detailing the construction of tools, including their names and descriptions. 
Next, we explain the simulation-based derivation of greedy and ground-truth trajectories using a tool graph framework, covering both non-blocking and blocking scenarios. 
We then present the evaluation metrics, including cost gaps, path dissimilarities, task completion rates, and tool call invalidity, along with their formulas and boundary conditions.
Finally, we describe the mechanisms for implementing blocking events, prove the solvability of blocking scenarios, and outline the randomized yet reproducible environment generation process to ensure fairness and reproducibility in evaluations.
An illustration of our environment is shown in Figure~\ref{fig:tool-flow-example}, while the primary hyperparameter configurations are presented in Table~\ref{tab:environment-hyperparas}.

\subsection{Tool Construction Details}
\label{app:tool-construction-details}

To better evaluate the model’s genuine planning ability, we mitigate potential overfitting to existing data~\cite{RL-data-contamination,MarCon}. Since biases acquired during training may distort the model’s assessment of tool cost-effectiveness, we construct a new set of tools for evaluation. In this section, we mainly introduce two aspects of the tool schema: name construction and description construction.

\paragraph{Tool Name Construction.}
The tool names in \cb{} are constructed systematically to reflect their function and position within the task hierarchy. This structured naming convention is designed to be both human-readable and machine-parsable, facilitating a clear understanding of each tool's purpose.

For \textbf{atomic tools}, names are generated based on a \texttt{Function\_Task\_Level} template. The \textit{Function} specifies the core action (e.g., \texttt{Decide}, \texttt{Search}, \texttt{Refine}, \texttt{Select}). The \textit{Task} denotes the domain-specific task (e.g., \texttt{Location}, \texttt{Transportation}). The \textit{Level} indicates the stage in the refinement process (e.g., \texttt{Preference}, \texttt{Step1}). An example is \texttt{Location\_Refinement\_Step1}, which clearly communicates its role.

For \textbf{composite tools}, which are sequences of atomic tools, names are generated algorithmically to summarize the encapsulated workflow. The naming logic depends on the composition of the toolchain. For instance, a two-step tool combining preference decision and initial search is named \texttt{[Task]\_Preference\_and\_Search}. A tool that covers the entire workflow from initial search to a specific refinement level is named \texttt{[Task]\_Search\_Planning\_to\_[Level]}. A complete pipeline from decision to final selection is named \texttt{[Task]\_Complete\_[N]Steps\_Pipeline}, where \texttt{N} is the number of atomic tools involved. This programmatic approach ensures that every possible contiguous sub-sequence of tools in a task plan has a unique and descriptive name.

\paragraph{Tool Description Construction.}
The descriptions for tools are crucial for the language model to understand their functionality and parameters. We employ a hybrid strategy for generating these descriptions, combining templated generation with LLM-based refinement to achieve both consistency and naturalness.

For \textbf{atomic tools}, descriptions are generated from a template that explains the tool's specific purpose. For example, the description for a refinement tool explicitly states the filtering dimension it applies, such as "Filtered by availability" for \texttt{Step1} or "Filtered by location" for \texttt{Step2}. This ensures that the model is aware of the distinct value each tool provides.

For \textbf{composite tools}, descriptions are generated using \textit{Gemini-2.5-Pro}~\cite{gemini}. We provide the LLM with the names and descriptions of the constituent atomic tools and prompt it to synthesize a concise, high-level summary of the overall workflow. For example, instead of merely concatenating the descriptions of a \texttt{Search} tool and a \texttt{Refine} tool, the LLM might generate a more intuitive description such as, ``Searches for options based on user preferences and then filters the results for availability''. This process is automated in our framework to generate descriptions for all possible composite tools.

It is worth noting that various aspects of tool information, such as cost, atomic or composite type, and the component tools of composite ones, are incorporated into the tool descriptions.

\begin{table}[t]
\centering
\small
\begin{tabular}{lc}
\toprule
Model & Average Tool Call Numbers \\
\midrule
\textit{Qwen3-8B} & 2.18 \\
\textit{Qwen3-14B} & 3.42 \\
\textit{Qwen3-32B} & 2.84 \\
\textit{Llama-3.1-8B-Instruct} & 1.99 \\
\textit{GLM-4.5} & 2.86 \\
\textit{Deepseek-V3.1} & 3.11 \\
\textit{Gemini-2.5-Pro} & 2.75 \\
\textit{Claude-Sonnet-4} & 2.62 \\
\textit{GPT-4o} & 3.36 \\
\textit{GPT-5} & 2.69 \\
\bottomrule
\end{tabular}
\caption{The statistics of average tool call counts for all evaluated models in the static evaluation (in Table~\ref{tab:main_unblocked_results}) include only valid tool calls; all invalid calls are excluded from the computation.}
\label{tab:tool_use_number_statistics}
\end{table}

\subsection{Greedy and Ground-truth Calculation}
\label{app:details-gt-greedy}

We emphasize that both the greedy and the ground-truth trajectories are obtained through \textit{simulation} rather than actual tool executions. To support this, we construct a \textbf{tool graph} that encodes the possible transitions between agent states.  

Formally, we define the tool graph $\mathcal{G} = (\mathcal{V}, \mathcal{E})$ as follows. Each vertex $v \in \mathcal{V}$ corresponds to an \textit{agent state}, represented as a set of datatypes already available to the agent:  
{
\begin{equation}
\small
v \equiv S = \{ d_1, d_2, \ldots, d_n \}, \quad S \subseteq \mathcal{D},
\end{equation}
}
where $\mathcal{D}$ is the universe of all possible datatypes. Importantly, states evolve \textit{monotonically}: invoking a tool never consumes existing datatypes but only augments the state with new ones, since the model always has access to the full conversation history.  

The initial state $S_0$ is determined by the task specification:  
{
\begin{equation}
\small
\begin{aligned}
S_0 = &\ \{\texttt{TimeInfo}\} 
       \cup \{\texttt{LocationPreference}\ \\ \text{(if needed)}\}
     & \cup \{\texttt{UserPreference}\}.
\end{aligned}
\end{equation}
}

For each task, we define a unique target datatype $\tau_{\text{task}}$, representing the final result required by the task, which is also the output of the last atomic tool.  
Accordingly, any state $S$ that contains $\tau_{\text{task}}$ is regarded as a goal state: 
{
\begin{equation}
\small
S \in \mathcal{V}, \quad \text{goal}(S) \iff \tau_{\text{task}} \in S.
\end{equation}
}

Each edge $e \in \mathcal{E}$ corresponds to the invocation of a tool. A tool $T$ is defined as a typed mapping  
{
\begin{equation}
\small
T: \mathcal{I}(T) \to \mathcal{O}(T),
\end{equation}
}
where $\mathcal{I}(T) \subseteq \mathcal{D}$ denotes the required input datatypes and $\mathcal{O}(T) \subseteq \mathcal{D}$ denotes the output datatypes. An edge exists from state $S$ to state $S'$ if and only if $\mathcal{I}(T) \subseteq S$, in which case  
{
\begin{equation}
S' = S \cup \mathcal{O}(T).
\end{equation}
}

Each edge is further assigned a \textit{cost} $c(T)$, representing the computational or interaction cost of invoking tool $T$. Thus, the transition can be written as  
{
\begin{equation}
(S \xrightarrow{\,T,\,c(T)\,} S') \in \mathcal{E}.
\end{equation}
}
This construction allows us to simulate greedy and ground-truth trajectories over the tool graph without actually executing tools. 
While such stimulation offers a convenient and controllable proxy, it fundamentally differs from real tool use: 
being algorithm-driven, it bypasses the core challenge of user-intent understanding. 
Consequently, although useful for isolating certain planning aspects, it fails to deal with the vagueness, ambiguity, and negotiation with human preferences inherent in real scenarios. 
Our benchmark therefore becomes essential: cost-aware planning cannot be addressed by traditional algorithmic simulation, but calls for evaluation of large language models, which uniquely possess the capacity to reason over ambiguity and align with human intent~\cite{tau-bench}.

\paragraph{Scenarios without Blocking.}

For scenarios without blocking, the construction of greedy and ground-truth trajectories is relatively straightforward.  
In the greedy policy, at each step given the current state $S$, we first compute the set of feasible tools:
{
\begin{equation}
\small
\mathcal{A}(S) = \{ T \mid \mathcal{I}(T) \subseteq S \ \wedge\ \mathcal{O}_{\text{prev}} \cap \mathcal{I}(T) \neq \varnothing \},
\end{equation}
}
where $\mathcal{I}(T)$ denotes the input types of tool $T$, and $\mathcal{O}_{\text{prev}}$ denotes the output type(s) of the previously invoked tool.  
For each candidate tool $T \in \mathcal{A}(S)$, we define a utility score that measures cost-efficiency:
{
\begin{equation}
\small
u(T) = \frac{c(T)}{\textit{comp}(T)},
\end{equation}
}
where $c(T)$ is the cost of invoking $T$, and $\textit{comp}(T)$ is the number of atomic tools contained in $T$.  
The greedy trajectory is then constructed by repeatedly selecting the tool with the smallest $u(T)$ until reaching a goal state.  
We explicitly restrict greedy to select only tools in $\mathcal{A}_{\text{chain}}(S)$ because any globally cost-optimal path must satisfy the chain property (i.e., each tool in the optimal path consumes the previous tool's output); by searching only within this subset of chain-consistent paths, the greedy policy explores exactly those paths that could possibly be globally optimal. 

In contrast, the ground-truth trajectory is obtained by directly running \textit{Dijkstra’s algorithm}~\cite{dijkstra} over the tool graph $\mathcal{G}$.  
Formally, let $\pi(S_0, S)$ denote the minimum-cost path from the initial state $S_0$ to a state $S$.  
The ground-truth trajectory is defined as
{
\begin{equation}
\small
\pi(S_0, S^\ast) = \arg\min_{\substack{S \in \mathcal{V} \\ \tau_{\text{task}} \in S}} \ \text{Cost}(S_0 \rightsquigarrow S),
\end{equation}
}
where $\text{Cost}(S_0 \rightsquigarrow S)$ is the cumulative cost of the path from $S_0$ to $S$.  
Unlike the greedy policy, this construction imposes no additional constraint that each tool must consume the immediate predecessor’s output, but instead corresponds to the standard shortest-path setting.

\paragraph{Scenarios with Blocking.}

For scenarios with blocking, trajectory construction is more involved due to unpredictable environment changes.  
Let $S_{t_i}$ denote the state immediately after the $i$-th blocking event, and 
$\mathcal{G}_{t_i}=(\mathcal{V}_{t_i},\mathcal{E}_{t_i})$ the updated tool graph at that time.  

For the greedy policy, the feasible tool set at state $S_{t_i}$ is

\begin{equation}
\small
\mathcal{A}^{(i)}(S_{t_i}) = 
\begin{cases}
    \{ T \mid \mathcal{I}(T) \subseteq S_{t_i} \}, & 
    \text{\parbox[t]{0.23\linewidth}{if this is the first step after blocking $i$,}} \\[6mm] 
    
    \{ T \mid \! \begin{aligned}[t] 
                & \mathcal{I}(T) \subseteq S_{t_i} \wedge \\
                & \mathcal{O}_{\text{prev}} \cap \mathcal{I}(T) \neq \varnothing 
             \end{aligned}
    \}, & 
    \text{otherwise.}
\end{cases}
\end{equation}

Here, the first-step relaxation occurs because the environment has changed, 
which effectively resets the greedy policy with a new initial state $S_{t_i}$.  
Greedy then selects
{
\begin{equation}
\small
T^\ast = \arg\min_{T \in \mathcal{A}^{(i)}(S_{t_i})} u(T), \quad 
u(T) = \frac{c(T)}{\textit{comp}(T)},
\end{equation}
}
and continues until reaching a goal state or the next blocking event.  
The tool graph $\mathcal{G}_{t_i}$ is updated according to the type of blocking: 
an edge is removed for a \emph{ban tool}, edge weights updated for a \emph{cost change}, 
unchanged for a \emph{preference change}, and accessible edges updated for a \emph{step length change}.  

For the ground-truth trajectory, we construct it in a segmented fashion. 
Let $S_{t_0} = S_0$ be the initial state.  
For each blocking interval $[t_i, t_{i+1})$, we compute the shortest path $\pi_i$ on the updated graph $\mathcal{G}_{t_i}$:
{
\begin{equation}
\small
\pi_i = 
\arg\min_{\substack{S \in \mathcal{V}_{t_i} \\ 
\tau_{\text{task}} \in S}} 
\text{Cost}_{\mathcal{G}_{t_i}}(S_{t_i} \rightsquigarrow S),
\end{equation}
}
where $\text{Cost}_{\mathcal{G}_{t_i}}(\cdot)$ denotes the cumulative cost in the updated graph.  

The full ground-truth trajectory is then obtained by concatenating all segments:
{
\begin{equation}
\small
\text{GT\_traj} = \pi_0 \oplus \pi_1 \oplus \cdots \oplus \pi_N,
\end{equation}
}
where $\oplus$ denotes path concatenation.  
This construction ensures that each segment is globally optimal under the current environment, 
while the full trajectory adapts to blocking events by replanning from the updated state each time.

\textbf{Ground-Truth Trajectory under Blocking:}
\begin{align*}
\small
&\text{Initial GT path:} \\
&S_0 \xrightarrow{A} S_1 \xrightarrow{B} S_2 \xrightarrow{C} S_3 \xrightarrow{D} S_4 \\
&\text{Blocking 1 after } B\text{, replan from } S_2: \\
&\pi_1: S_2 \xrightarrow{E} S_5 \xrightarrow{F} S_6 \xrightarrow{G} S_7 \\
&\text{Blocking 2 after } F\text{, replan from } S_6: \\
&\pi_2: S_6 \xrightarrow{H} S_8 \xrightarrow{I} S_9 \\
&\text{Full GT trajectory:} \\
&\text{GT\_traj} = \pi_0 \oplus \pi_1 \oplus \pi_2
\end{align*}

Above is an example of a ground-truth trajectory under blocking. Each blocking event triggers replanning from the current state. Segments $\pi_i$ correspond to shortest paths computed on the updated tool graph after the $i$-th blocking event. The $\oplus$ operator denotes concatenation of path segments.

Note that while our randomized cost assignment strategy naturally minimizes the likelihood of identical path costs, ties may still theoretically occur. 
To ensure a unique ground truth, we implement a tie-breaking rule: 
among paths with equivalent total costs, the trajectory with fewer tool calls is selected as the optimal solution.
This criterion is strictly enforced in our evaluation and is also explicitly communicated to the models via the inference prompt (see Figure~\ref{fig:runtime-prompt}).

\begin{figure*}
\centering

\includegraphics[width=\linewidth]{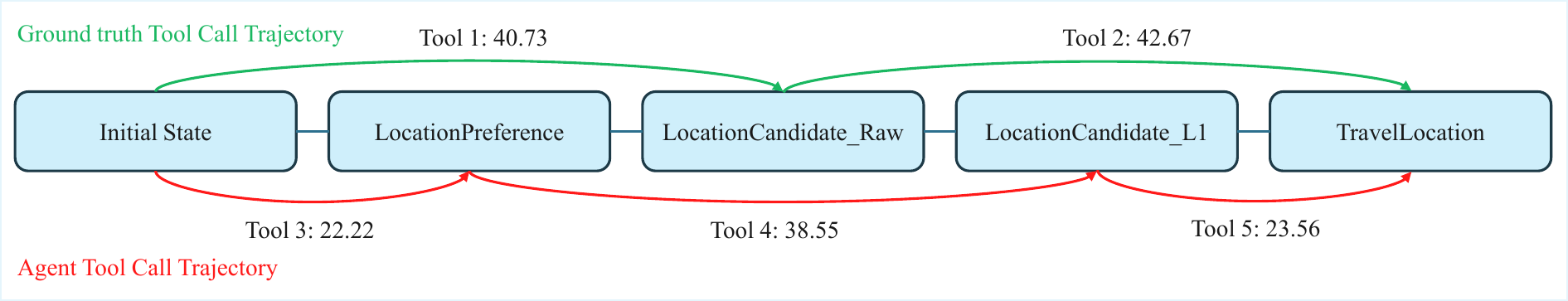}
\caption{An illustrative example of metric calculation. The diagram contrasts the ground-truth trajectory ($\tau_{gt}$, indicated in \textcolor{deepgreen}{green}) with the agent's trajectory ($\tau_{agent}$, indicated in \textcolor{deepred}{red}). The numbers on the edges represent the specific cost of each tool invocation. We use this scenario to demonstrate how Cost Gap, path-related metrics (AED, ANED, EMR), and task completion metrics are computed.}
\label{fig:cost-gap-example}
\end{figure*}

\subsection{Metric Details}
\label{app:detailed-metrics}

\subsubsection{Detailed Implementation and Formulas}
\label{app:metrics-implementation}

\paragraph{Cost Metrics.} 
\label{app:cost-metrics-explanation}
The design intuition of the Cost Gap metric is to directly capture how much more (or less) cost an agent incurs compared to the ground-truth trajectory.
Unlike absolute cost values, which depend on the specific scale of the tool graph and are not directly comparable across different settings, the gap formulation provides a simple, interpretable measure of deviation from the optimal reference.

Formally, let $\tau_{\text{agent}} = (a_1, a_2, \ldots, a_m)$ and $\tau_{\text{gt}} = (g_1, g_2, \ldots, g_n)$ denote the sequences of actions in the agent and ground-truth trajectories, respectively. Each action $x$ is associated with a cost $\text{cost}(x)$.
The cost gap is defined as:
{
\begin{equation}
\small
\text{Cost Gap} = 
\sum_{a \in \tau_{\text{agent}}} \text{cost}(a)
- \sum_{g \in \tau_{\text{gt}}} \text{cost}(g).
\end{equation}
}

By construction, $\text{CostGap} \geq 0$ since the agent is less cost-efficient than the ground-truth in the static setting, and $\text{CostGap} = 0$ indicates perfect alignment with the ground-truth trajectory.

In Table~\ref{tab:main_unblocked_results}, we present two variants of the cost gap metric. The values outside the parentheses denote the average cost gap computed over all samples, whereas the values in parentheses are obtained after excluding two types of progress-awareness failures: (i) extra tool calls, where the agent continues invoking tools after reaching the goal state, and (ii) repeated tool calls, where the agent redundantly invokes the same tool multiple times even after the initial invocation has already succeeded. Our intuition is to decouple progress-awareness from cost-awareness, since extra or repeated calls primarily indicate a lack of progress-awareness rather than deficiencies in cost reasoning. Detailed statistics of such progress-aware errors are reported in Table~\ref{tab:error-analysis-statistics}.

Note that we do not compare the aggregated cost optimality over the entire trajectory under blocking scenarios. 
This is because, at the initial state or immediately after each blocking event, the planning procedure assumes the current goal state is the true final goal. 
As a result, the cumulative trajectory cost under blocking does not reliably capture cost optimality, since a lower total cost does not necessarily correspond to a better trajectory.

Another important aspect is that, by design, the environment assigns each composite tool a cost equal to the sum of its component atomic tools plus zero-mean Gaussian noise. As a result, the expected cost of any tool-call sequence reaching the goal state is identical, regardless of its length. Therefore, the cost gap cannot be reduced to a proxy for the number of tool calls; instead, it captures whether the model adopts cost-optimal invocation patterns under the given cost structure.

\paragraph{Path Metrics.} 
For \textbf{Average Edit Distance (AED)}, we use an extended definition of the \textit{Levenshtein Distance}~\cite{edit-distance, l_distance}, defined as the minimum number of operations required to transform one path into another. Here, each path is represented as a sequence of tool calls, and the allowed operations are insertion, deletion, and substitution of tool calls. This metric directly measures the structural dissimilarity between the agent's trajectory and the ground-truth trajectory. 
{
\begin{equation}
\small
\text{ED}(\tau_{\text{agent}}, \tau_{\text{gt}}) = \min_{\Gamma \in \mathcal{O}(\tau_{\text{agent}} \rightarrow \tau_{\text{gt}})} |\Gamma|,
\end{equation}
}
where $\mathcal{O}(\tau_{\text{agent}} \rightarrow \tau_{\text{gt}})$ denotes the set of valid edit-operation sequences transforming $\tau_{\text{agent}}$ into $\tau_{\text{gt}}$, and $|\Gamma|$ is the number of operations in $\Gamma$. 
We compute the Average Edit Distance across all queries in the test set as:
{
\begin{equation}
\small
\text{AED} = \frac{1}{N} \sum_{i=1}^{N} \text{ED}(\tau_{\text{agent}}^{(i)}, \tau_{\text{gt}}^{(i)}),
\end{equation}
}
where $N$ is the total number of queries in the test set.

For \textbf{Average Normalized Edit Distance (ANED)}, we follow~\citet{norm-edit-distance} and normalize the edit distance by the maximum of the two path lengths. For each query, the Normalized Edit Distance (NED) is defined as:
{
\begin{equation}
\small
\text{NED}(\tau_{\text{agent}}, \tau_{\text{gt}}) = \frac{\text{ED}(\tau_{\text{agent}}, \tau_{\text{gt}})}{\max\big(|\tau_{\text{agent}}|, \, |\tau_{\text{gt}}|\big)},
\end{equation}
}
where $|\tau|$ denotes the length of a trajectory. By construction, this metric ranges from $0$ to $1$, where $0$ indicates identical paths and $1$ indicates maximal dissimilarity. We then average across all queries:
{
\begin{equation}
\small
\text{ANED} = \frac{1}{N} \sum_{i=1}^{N} \text{NED}(\tau_{\text{agent}}^{(i)}, \tau_{\text{gt}}^{(i)}).
\end{equation}
}

For \textbf{Exact Match Ratio (EMR)}, we use a binary indicator that checks whether the agent's trajectory exactly matches the ground-truth trajectory. For each query, the exact match is defined as:
{
\begin{equation}
\small
\text{EM}(\tau_{\text{agent}}, \tau_{\text{gt}}) = \mathbb{1}\!\left[\tau_{\text{agent}} = \tau_{\text{gt}}\right],
\end{equation}
}
where $\mathbb{1}[\cdot]$ is the indicator function that equals $1$ if the two trajectories are identical and $0$ otherwise. The Exact Match Ratio is then computed as the proportion of exact matches across the test set:
{
\begin{equation}
\small
\text{EMR} = \frac{1}{N} \sum_{i=1}^{N} \text{EM}(\tau_{\text{agent}}^{(i)}, \tau_{\text{gt}}^{(i)}).
\end{equation}
}

\paragraph{Task Completion Metric.} 
We report \textbf{Task Completion Ratio} as the proportion of samples where the agent’s predicted final candidate exactly matches the ground-truth answer. 
This mainly captures the agent’s ability to understand and interpret human intent expressed in natural language, as well as to invoke the necessary tools to obtain the final answer.

\paragraph{Tool Call Metric.}

%
For the \textbf{Invalid Tool-Use Ratio}, we report the proportion of invalid tool calls over all tool calls.

\subsubsection{Boundary Conditions}
\label{app:metrics-boundary-conditions}

\paragraph{Cost Metrics.} 
For cost-related metrics, we exclude all data points where the agent fails to reach the goal state within the maximum of 20 turns, since comparing a trajectory that stops at an intermediate state with one that reaches the goal is not meaningful. 
When computing the total cost of each data point, we also disregard invalid search attempts. 
This design ensures that the reported cost reflects the agent’s genuine cost awareness, while disentangling it from its tool-use capability.

\paragraph{Path Metrics.} 
For path-related metrics, we focus on evaluating genuine cost-optimal planning ability. 
Trajectories that fail to reach the goal state are excluded from metric computation to ensure that cost awareness is measured only on successful completions.
To avoid being influenced by agents' tool calling ability and reflect true cost-optimal planning capability, we also exclude failed tool calls. 
We thereby disentangle agents' tool calling ability and cost-optimal planning ability.

\paragraph{Task Completion metric.}

For \textbf{Task Completion Ratio}, we exclude all samples in which the agent fails to reach the goal state, since such predictions amount to mere guesses rather than answers derived from a complete tool-call trajectory. 

\paragraph{Tool Call Metrics.}

For the \textbf{Invalid Tool-Use Ratio}, we include all data points regardless of outcome. 
This metric reflects the model’s zero-shot tool-use ability on our benchmark, providing an intuition of how such ability impacts cost-optimal planning performance.

Notably, in blocking scenarios, we exclude all samples that do not experience the designated number of blocking events.
Including such samples would make the metrics unfair; therefore, we report results based on the intersection of valid sets that meet the blocking criteria.

\subsubsection{Illustrative Examples}
\label{app:metrics-example}

To better understand the proposed metrics, we present a concrete example based on the trajectories shown in Figure~\ref{fig:cost-gap-example}. 
Let $\tau_{gt}$ denote the Ground-truth trajectory (green path), which employs a sequence of two composite tools with a total cost of $40.73 + 42.67 = 83.40$. 
In contrast, let $\tau_{agent}$ denote the Agent trajectory (red path), which adopts a less efficient plan consisting of three tools with a total cost of $22.22 + 38.55 + 23.56 = 84.33$.
In the following, we illustrate our proposed metrics one by one using this example.

\noindent \textbf{(1) Cost Gap.} 
This metric quantifies the economic inefficiency of the agent. 
In this scenario, the Cost Gap is calculated as the difference between the agent's total cost and the ground-truth cost: $84.33 - 83.40 = 0.93$. 
This positive value indicates that the agent failed to find the most cost-effective solution compared to $\tau_{gt}$.

\noindent \textbf{(2) Path Metrics (EMR, AED, ANED).} 
These metrics assess the structural alignment between $\tau_{agent}$ and $\tau_{gt}$ . 
Specifically, the \textbf{Exact Match Ratio (EMR)} is 0 for this instance because the agent's tool sequence ($\tau_{agent}$) differs markedly from the ground-truth sequence ($\tau_{gt}$). 
The \textbf{Average Edit Distance (AED)} measures the minimum number of edit operations required to transform $\tau_{agent}$ into $\tau_{gt}$. 
In this specific case, since the agent's tool chain (Tool 3 $\to$ Tool 4 $\to$ Tool 5) is completely disjoint from the ground truth (Tool 1 $\to$ Tool 2), the transformation necessitates three operations.
For instance, substituting Tool 3 with Tool 1, substituting Tool 4 with Tool 2, and deleting Tool 5 results in an edit distance of 3.
The \textbf{Average Normalized Edit Distance (ANED)} normalizes the AED by the maximum length of the two trajectories (i.e., $\max(|\tau_{agent}|, |\tau_{gt}|) = \max(3, 2) = 3$).
This yields an ANED of $3/3 = 1.0$, quantifying that the agent's plan exhibits a 100\% structural divergence from the optimal strategy.
Note that the final reported values for these metrics are computed as the mean across all test instances.

\noindent \textbf{(3) Task Completion Ratio (TCR).} 
Despite the sub-optimal path, the agent successfully reached the final goal state (\textit{TravelLocation}). 
Therefore, the TCR for this sample is 1 (or 100\%), highlighting that the agent was effective in task completion but lacked efficiency.
The final reported TCR is calculated as the average over all samples.

\noindent \textbf{(4) Invalid Tool-Use Ratio (ITUR).} 
This metric measures the proportion of failed tool invocations, such as calling non-existent tools, using wrong parameters, or violating prerequisites. 
In the example of Figure~\ref{fig:cost-gap-example}, all tool calls executed by the agent were valid transitions within the tool graph, resulting in an ITUR of 0.
However, in cases where an agent attempts to invoke a tool without satisfying its input dependencies (e.g., calling a tool that requires \textit{LocationCandidate\_L1} without first obtaining it), such calls would be counted as invalid failures.

\subsection{Blocking Details}
\label{app:blocking-details}

\paragraph{Block Trigger Time.}
To ensure blocking events occur at strategically meaningful moments throughout the agent's planning process, we employ a dynamic trigger mechanism that adapts to the evolving optimal path length after each environment change.
Specifically, let $L_i$ denote the optimal path length after the $i$-th blocking, $t_i$ denote the current execution step, and $B$ denote the total number of planned blockings.
For the next ($i+1$)-th blocking, we calculate the relative trigger position as $\Delta = \lfloor L_i / (B - i) \rfloor$, and set the trigger step to $t_{i+1} = t_i + \max(1, \Delta)$.
This strategy distributes blocking events approximately evenly across the remaining optimal trajectory, preventing clustering at the beginning or end of execution.
Critically, the trigger step for each blocking is recalculated dynamically after every environment change, as each blocking (especially cost changes or tool removals) may alter the optimal path length $L_i$.

\subsection{Proof of Solvability of Blocking Scenarios}
\label{app:block-but-solvable}

For four block types: ban tool, preference change, cost change and remove tools, we both ensure that the agent could reach the goal state in theory, that is to say, the tool graph remains solvable.

\begin{itemize}
    \item \textbf{Ban Tool} Intuitively, the worst case occurs when all blockings prevent the same state from transitioning to other states. 
    As long as the agent can still move to alternative states, it can reach the goal. 
    Therefore, when the tool that completes the entire task in a single step is unavailable, the maximum number of blockings that can be tolerated is $task\_length - 2$. 
    A rigorous proof is provided later in this section.

    \item \textbf{Preference Change} Since the tool graph does not change, the solvability is unaffected.
    \item \textbf{Cost Change} It is trivial that the tool graph could still be solved, since no edges are removed from the graph.
    \item \textbf{Remove Tools} In this case, we always keep the atomic tools unremoved to maintain solvability.
\end{itemize}

Rigorous proof for the ``Ban Tool'' scenario:
    Consider a linear task of length \(n\). Suppose every contiguous interval \([i,j]\) with \(1\le i\le j\le n\) corresponds to a tool, except that the full-length tool \([1,n]\) is banned by default. Then the minimum number of additional tools whose removal destroys \emph{all} partitions of \([1,n]\) is
    \[
    h = n-1.
    \]
    Consequently, the robustness bound is
    {
    \[
    K_{\mathrm{robust}} = h-1 = n-2,
    \]
    }
    meaning that as long as fewer than \(n-1\) tools are banned, the tool graph always remains solvable.

    For each \(k\in\{1,\dots,n-1\}\), consider the partition
    {
    \[
    P_k = \{[1,k],[k+1,n]\}.
    \]
    }
    Since \([1,n]\) is not allowed, each \(P_k\) is a valid partition. Distinct \(k\) yield disjoint sets of tools, so to block all partitions one must remove at least one tool from each \(P_k\), i.e.\ at least \(n-1\) tools. Thus \(h\ge n-1\). 

    Conversely, let 
    {
    \[
    H=\{[1,1],[1,2],\dots,[1,n-1]\}.
    \]
    }
    Any partition of \([1,n]\) must begin with some \([1,k]\) (\(1\le k\le n-1\)), because \([1,n]\) is banned. Hence every partition uses a tool from \(H\). Therefore \(H\) is a hitting set of size \(n-1\), so \(h\le n-1\). Combining gives \(h=n-1\), and thus the robustness bound is \(K_{\mathrm{robust}}=n-2\), which completes the proof.

\subsection{Randomized yet Reproducible Environment}
\label{app:randomized-yet-reproducible-environment}
To enable systematic evaluation while maintaining experimental variability, we design a fully deterministic environment generation framework controlled by a global seed $S$. This ensures that all random aspects (tool costs, blocking parameters, and trigger timings) are reproducible across runs while remaining sufficiently diverse across different queries.
\subsubsection{Deterministic Cost Assignment}
\label{app:cost-assignment}
For each query with identifier $q$, and for each tool (atomic or composite) with name $\text{name}(\cdot)$, we deterministically derive random seeds by applying SHA-256 hashing~\cite{nist-sha-256} to the tuple $(S, q, \text{name})$:
{
\begin{gather}
\small
c_{q,a} \sim \operatorname{Uniform}(c_{\min}, c_{\max}; \text{seed}=h(S,q,a)) \tag{1} \\
\epsilon_{q,T} \sim \mathcal{N}(0,\sigma^2 k; \text{seed}=h(S,q,T)) \tag{2} \\
\mbox{\footnotesize$\displaystyle
C_{q,T} = \max\left(1.00, \operatorname{round}\left(\sum_{a\in T} c_{q,a} + \epsilon_{q,T}, 2\right)\right)
$} \tag{3}
\end{gather}
}
Here $a$ denotes an atomic tool, $T$ a composite tool with $k$ atomic components, and $h(\cdot)$ the hash-based seed derivation function. The $\sqrt{k}$ scaling in equation (2) reflects empirical observations that execution variance grows sublinearly with pipeline length~\cite{Theory-of-Scheduling, Factory-Physics}. 

\subsubsection{Deterministic Blocking Parameters}
\label{app:deterministic-blocking-parameters}

When evaluating agents under a specific blocking type (ban\_tool, preference\_change, remove\_tools, or cost\_change), all blocking events are pre-planned using a hierarchical seeding strategy. Given a base seed $S_q$ for query $q$, we derive the seed for the $i$-th blocking as $S_{q,i} = S_q + i \cdot \Delta_s$, where $\Delta_s = 100$ is a fixed interval ensuring seed independence. This seed $S_{q,i}$ deterministically controls the parameters for blocking $i$:
\begin{itemize}[itemsep=0pt,topsep=2pt]
\item \textbf{Cost Change:} A new global seed $S'$ is sampled from a predefined range using $S_{q,i}$, which then regenerates all tool costs via equations (1)–(3).
\item \textbf{Preference Change:} $S_{q,i}$ selects an alternative query from the same task category to extract new user requirements and preferences.
\item \textbf{Remove Tools:} 
Using $S_q$, we deterministically sample $n$ non-overlapping length intervals $[l_i, r_i]$ from the feasible range $[2, L_{\text{max}}]$, where $n$ is the total count of \textit{remove tools} events and $L_{\text{max}}$ depends on the refinement level. 
As shorter tools are more prevalent, we impose a minimum width constraint $(r_i - l_i) \geq L_{\text{max}}/2$ to ensure the removal of a sufficient number of tools, thereby increasing the likelihood of modifying the ground-truth plan.
At blocking $i$, only composite tools with length exactly equal to $l_i$ (the left endpoint) are removed from the visible tool set, while all atomic tools remain accessible.
\item \textbf{Ban Tool:} $S_{q,i}$ produces deterministic failure messages, where the banned tool is determined at runtime according to the agent’s actual tool call, thereby enabling targeted blocking of the agent’s actions.
\end{itemize}

Thus, this hierarchical seeding ensures that: (i) given $(S, q)$ and the blocking type, the entire blocking sequence and all associated parameters are fully determined (only the trigger time would be different, and we will explain its fairness in Appendix~\ref{app:blocking-fairness}); (ii) different queries exhibit different blocking patterns with the same $S$; and (iii) experimental results are perfectly reproducible, facilitating controlled comparisons across models and further analysis. 

\begin{table*}[htbp]
\centering
\small
\begin{tabular}{lcccc}
\toprule
\multicolumn{1}{c}{\multirow{1}{*}{\textbf{Steps}}} & \multicolumn{2}{c}{\textbf{Redundant Tool calls}} & \multicolumn{2}{c}{\textbf{Failure Tool Calls}} \\
        \cmidrule(lr){2-3} \cmidrule(lr){4-5}
        & Repeated Calls $\downarrow$ & Extra Calls $\downarrow$ &  Wrong Parameters $\downarrow$ & Inaccessible Calls $\downarrow$ \\
\midrule
\textit{Qwen3-8B} & 1  & 0  & 0  & 6 \\
\textit{Qwen3-14B} & 1  & 7 & 1 & 11  \\
\textit{Qwen3-32B} & 0 & 0 & 19  & 0 \\
\textit{Llama-3.1-8B-Instruct} & 0 & 0 & 19 & 347 \\
\textit{GLM-4.5} & 2 & 0 & 3 & 43 \\
\textit{Deepseek-V3.1} & 16 & 0 & 6 & 314\\
\textit{Gemini-2.5-pro} & 0 & 2 & 0 & 0 \\
\textit{Claude-sonnet-4} & 3 & 0 & 0 & 53 \\
\textit{GPT-4o} & 10 & 0 & 1 & 318 \\
\textit{GPT-5} & 11 & 0 & 0 & 1  \\
\bottomrule
\end{tabular}
\caption{Detailed statistics of failure modes for the tested models in the main experiment (Table~\ref{tab:main_unblocked_results}). Numbers in parentheses represent the ratio (\%) to the total number of tool calls.}
\label{tab:error-analysis-statistics}
\end{table*}

\subsubsection{Evaluation Fairness}
\label{app:blocking-fairness}

\paragraph{Dynamic blocking mechanism ensures fairness across models.} Among the four types of blocking events, all trigger times are determined based on the optimal path. 
Therefore, given identical cost assignments, the first trigger time remains fixed.
However, the subsequent blocking time depends on the model's behavior, because the state the model reaches during the first block determines the optimal path and thus the trigger time for the next blocking. We argue this is fair across models:

\begin{itemize}
    \item If a model follows the correct trajectory, both the ground truth and all agents are identically affected during the whole interaction trajectory, preserving fairness in evaluation.
    \item Conversely, if a model already deviates from the optimal path, the model-dependent blocking trigger time has no impact, as the data point’s EM score would already be zero. Consequently, the \textbf{EMR} metric remains unaffected.
    \item Moreover, as shown in Table~\ref{tab:main_unblocked_results}, \textbf{EMR} and \textbf{ANED}/\textbf{AED} exhibit opposite trends, suggesting that the latter two effectively capture performance differences as well.
\end{itemize}

Moreover, the blocking parameters are determined through a pseudo-random seeding mechanism (see details in Appendix~\ref{app:deterministic-blocking-parameters}), making them deterministic and consistent across all models. Therefore, we consider our dynamic blocking mechanism to be fair across models.

\subsubsection{Evaluation Robustness}

\paragraph{Random seed mechanism promotes evaluation robustness.}

Building on the deterministic seeding framework introduced in Appendix \ref{app:deterministic-blocking-parameters}, our evaluation benefits from the diverse cost assignment and thereby achieves robustness. 
Specifically, the global seed $S$ is hierarchically propagated into per-query seeds $S_q$ and subsequently into tool-level seeds through the hash-based function $h(S,q,\text{name})$ (Eq. 1–3). This mechanism ensures that each query instantiates a distinct cost landscape: 
atomic tools receive independently sampled costs, and composite tools derive their perturbed costs via $\epsilon_{q,T}$, which yield sufficiently varied environments that reveal the agent’s true cost-awareness rather than memorization of any fixed structure.
Additionally, we evaluate models under three independent runs (details in Appendix~\ref{app:effect-of-random-seeds}), each initialized with a different global seed $S$. The consistency of results across these runs confirms that the benchmark remains robust to seed variation, while the hierarchical generation preserves controlled diversity across instances.

\paragraph{Deterministic blocking content poses no memorization risk.}

Although the specific blocked tools and new cost values are pre-determined by a fixed seed for reproducibility, they remain completely invisible to the agent until runtime, and our evaluation architecture explicitly forbids any cross-instance memory or episodic recall. 
Each task is executed in full isolation with a fresh context, making pattern exploitation impossible.


\subsubsection{Discussions on Evaluation}

\paragraph{The necessity of introducing controlled uncertainty.} 

While fully deterministic blocking (i.e., identical blocking parameters and trigger times) would benefit cross-model comparability, it would fundamentally limit the benchmark's ability to evaluate adaptive planning. 
Real-world environments are inherently dynamic—cost structures shift, tools fail, and preferences change—requiring agents to replan under uncertainty rather than memorize fixed trajectories. 
If blocking events were static across all queries, a model could trivially overfit to CostBench through fine-tuning or pattern memorization, achieving superior scores without improving its robustness or generalization, as observed in prior benchmarks prone to leakage~\cite{RL-data-contamination, RFMBench}. 
Introducing controlled uncertainty addresses this issue: the content of each blocking event is deterministic under a global seed, ensuring reproducibility, while the timing adapts to the agent’s progress, preventing shortcut exploitation and preserving evaluation diversity. This hybrid design strikes a balance—maintaining fairness through seed-controlled reproducibility, while providing enough diversity to meaningfully assess an agent’s ability to detect disruptions, revise plans, and recover from unexpected environmental changes.

\section{Additional Experiment Results}

\subsection{Hardest Test on Static Settings}
\label{app:task-sequence-8-analysis}

In Figure~\ref{fig:unblocked_scaling_combined}, we analyze model performance in static settings, with task sequence lengths ranging from 5 to 8. We further report the detailed performance values in Table~\ref{tab:more_unblocked_results} for the most challenging case, task sequence = 8. We observe that under this setting, the EMR of all Qwen-series models drops to single digits. \textit{Gemini-2.5-Pro} also exhibits a substantial degradation, falling from 83.73 at task sequence = 5 to 35.15, while only \textit{GPT-5} maintains an EMR of approximately 75\%. These results indicate that current models struggle even with relatively simple forms of complexity scaling. Notably, we argue that comparisons among Qwen-series models in this setting are of limited practical significance, as their performance is nearly indistinguishable from a greedy policy, suggesting that their decisions are effectively equivalent to random guessing.

\subsection{Evaluation Robustness}
\label{app:effect-of-random-seeds}

\begin{table}[htbp]
\small
\resizebox{\columnwidth}{!}{%
\begin{tabular}{lccc}
\toprule
\textbf{Models} & \makecell{\textbf{Radius} \\ \textbf{(EMR)}} (\%) & \makecell{\textbf{Radius} \\ \textbf{(ANED)}} (\%) & \makecell{\textbf{Radius} \\ \textbf{(AED)}} \\
\midrule
\textit{Qwen3-8B}              & 2.82 & 2.55 & 0.085    \\
\textit{Qwen3-14B}             & 3.59 & 2.83 & 0.132    \\
\textit{Qwen3-32B}             & 3.72 & 3.10 & 0.120    \\
\textit{Llama-3.1-8B-Instruct} & 2.02 & 2.07 & 0.112    \\
\textit{GLM-4.5}               & 3.04 & 2.94 & 0.101    \\
\textit{Deepseek-V3.1}         & 2.82 & 2.51 & 0.117    \\
\textit{Gemini-2.5-Pro}        & 2.81 & 1.64 & 0.067    \\
\textit{Claude-Sonnet-4}     & 3.86 & 2.96 & 0.105    \\
\textit{GPT-4o}                & 2.55 & 2.23 & 0.110    \\
\textit{GPT-5}                 & 1.11 & 1.56 & 0.047    \\
\hline
\textbf{\textit{Average}}      & 2.83 & 2.44 & 0.100     \\
\bottomrule
\end{tabular}%
}
\caption{95\% Confidence Intervals for Main Metrics. We employ the Bootstrap method~\cite{Boostrap-confidence-intervals} with 10,000 resampling iterations on the test set ($N=381$) to estimate statistical uncertainty. The reported \textbf{Radius} values denote the half-width of the 95\% confidence interval (i.e., the $\pm$ value) for Exact Match Ratio (EMR), Average Normalized Edit Distance (ANED), and Average Edit Distance (AED). }
\label{tab:confidence-interval}
\vspace{-0.2in}
\end{table}

\paragraph{Theoretical Bounds.}

To ensure the statistical validity of our evaluation, we quantified the uncertainty associated with our metrics using the Bootstrap method~\cite{Boostrap-confidence-intervals} with $10,000$ resampling iterations. 
As presented in Table~\ref{tab:confidence-interval}, our measurement exhibits relatively high precision across all evaluated models. 
The 95\% confidence intervals are remarkably tight, with an average half-width radius of only $2.83\%$ for Exact Match Ratio (EMR) and $2.44\%$ for Average Normalized Edit Distance (ANED). 
Furthermore, the radius for Average Edit Distance (AED) is negligible at $0.100$. 
These narrow bounds confirm that the observed performance differences between models are statistically significant and not artifacts of sampling noise, demonstrating the robustness of our benchmarking methodology.

\paragraph{Empirical Results.}

\begin{figure}
    \centering
    \includegraphics[width=\linewidth]{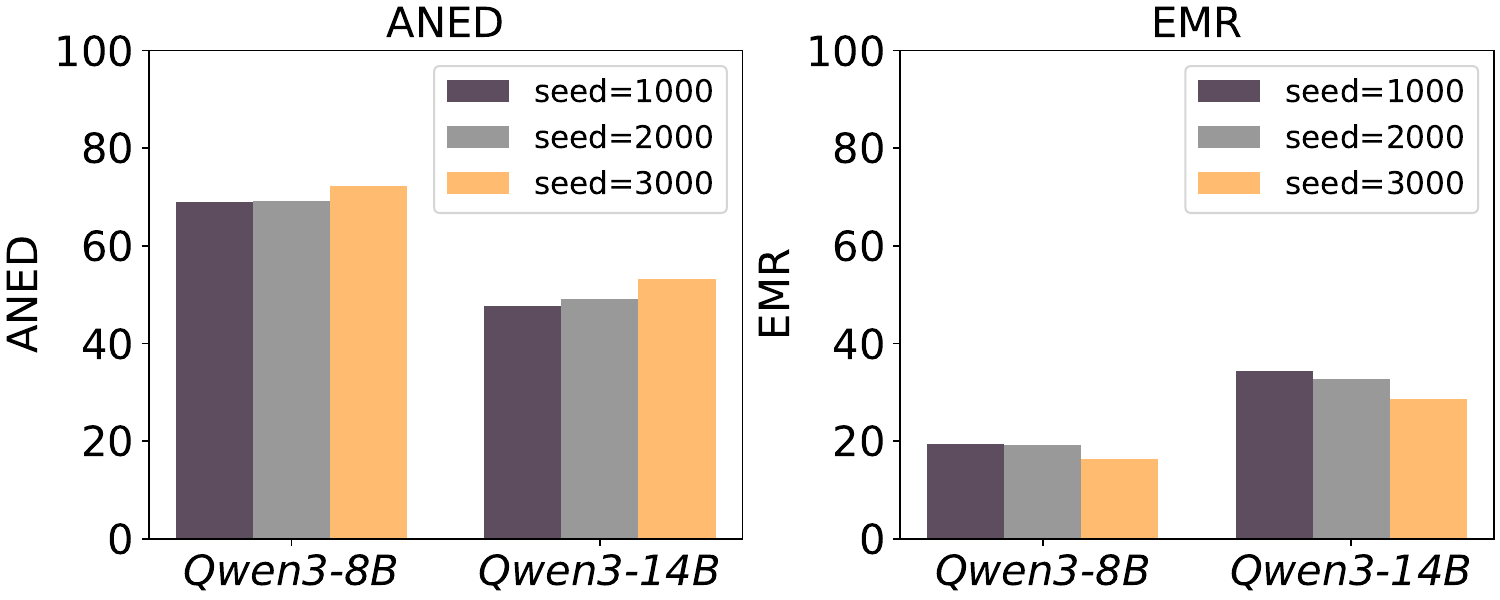}
    \caption{Performance of Qwen3-8B and Qwen3-14B on ANED and EMR \textbf{(\%)} across different seeds. Results demonstrate low variations with different seeds, with variations in ANED and EMR not exceeding 5\% across seeds.}
    \label{fig:seed-ablation}
    \vspace{-2mm}
\end{figure}

To further empirically ensure the robustness of our conclusions, which is controlled by a random seed, we conducted three additional experiments using seeds 1000, 2000, and 3000 to evaluate model performance. 
As illustrated in Figure~\ref{fig:seed-ablation}, the LLMs exhibit no significant performance variation across different seeds, with differences in ANED and EMR remaining within 5\%, confirming low seed sensitivity. 

\section{Analysis and Discussion}

\begin{table*}[t]
    \centering
    \scriptsize
    \resizebox{\linewidth}{!}{
    \begin{tabular}{l c c c c}
        \toprule
        \multicolumn{1}{c}{\multirow{1}{*}{\textbf{Steps}}} & \multicolumn{2}{c}{\textbf{Redundant Tool calls}} & \multicolumn{2}{c}{\textbf{Failure Tool Calls}} \\
        \cmidrule(lr){2-3} \cmidrule(lr){4-5}
        & Repeated Calls & Extra Calls &  Wrong Parameters & Inaccessible Calls \\
        \midrule
        Step 1 & \scriptsize{Decide\_Attraction\_Preference} & \scriptsize{Decide\_Shopping\_Preference} & \scriptsize{Attraction\_Full\_Planning\_to\_Step1} & \scriptsize{Decide\_Location\_Preference} \\
        Step 2 & \textcolor{deepred}{\scriptsize{Attraction\_Preference\_and\_Search}} & \scriptsize{Search\_Shopping\_Candidates} & \textcolor{deepred}{\scriptsize{Attraction\_Finalize\_from\_Step1\_3Steps}} & \scriptsize{Search\_Location\_Candidates} \\
        Step 3 & \scriptsize{Attraction\_Refine\_to\_Step2} & \textcolor{deepgreen}{\scriptsize{Shopping\_Finish\_from\_Step1\_3Steps}} & - & \textcolor{deepred}{\scriptsize{Location\_Refinement\_Step2}} \\
        Step 4 & \textcolor{deepgreen}
{\scriptsize{Select\_Final\_Attraction}} & \textcolor{deepred}{\scriptsize{Shopping\_Refinement\_Step1}} & - & - \\
        Step 5 & - & \textcolor{deepred}{\scriptsize{Shopping\_Refinement\_Step2}} & - & - \\
        Step 6 & - & \textcolor{deepred}{\scriptsize{Select\_Final\_Shopping}} & - & - \\
        \bottomrule
    \end{tabular}}
    \caption{Representative examples of the four failure modes, taken from \textit{Qwen3-14B}. Tool calls highlighted in \textcolor{deepgreen}{green} indicate the step where the goal state is reached, while those in \textcolor{deepred}{red} mark the erroneous calls. Each model in Table~\ref{tab:main_unblocked_results} exhibits at least one of these error types.}
    \label{tab:error-analysis-cases}
\end{table*}

\subsection{Data Distribution}
\label{app:data-distribution}

We note a slight imbalance in our final filtered dataset, where the ``Location'' task contains the fewest instances in both the training and test splits. 
This skew originates from our generation pipeline: we initially sampled an equal number of raw preference combinations for each task, but a subsequent commonsense filter retained a different proportion of combinations for each task. 
The ``Location'' task had a lower filter pass rate due to its unique dimensional features, resulting in the final distribution. 

We argue that this imbalance is non-critical for our primary objective of evaluating cost-optimal planning. 
The user preference features are only utilized in the initial step to understand user intent and select the first tool. 
The subsequent agent trajectory, which comprises the core planning and tool-calling sequence, is independent of these initial preferences. 
Therefore, while the skew may slightly influence the evaluation of intent understanding, it does not affect the integrity of the cost-optimal planning assessment.
For future applications, users of \cb{} requiring a more balanced distribution or augmented ``Location'' data can readily achieve this by adjusting parameters in our provided codebase.

\subsection{Error Analysis}
\label{app:error-analysis}

Building upon the main discussion in Section~\ref{sec:analysis}, we provide a detailed examination of all observed error types, including both redundant and failure tool calls. These analyses shed light on how progress-awareness limitations hinder the agents’ ability to plan and act cost-efficiently.

\subsubsection{Redundant Calls}

As defined in Section~\ref{sec:analysis}, redundant calls consist of two subtypes:  
(1) \textit{Repeated calls}, where the model invokes the same tool multiple times even after a successful call, and  
(2) \textit{Extra calls}, where the model continues invoking tools after reaching the goal state.  

Such behaviors commonly occur when models fail to maintain an internal notion of task completion or overlook that an equivalent operation has already been performed.  
In Table~\ref{tab:error-analysis-statistics}, we report the frequency of these redundant patterns across models. The overall ratio of redundant calls correlates with the noise observed in the raw cost metrics, as discussed in Section~\ref{sec:analysis}.

Table~\ref{tab:error-analysis-cases} illustrates representative examples:
\begin{itemize}
    \item \textbf{Repeated Calls:} The agent redundantly invokes both ``Decide\_Attraction\_Preference'' and ``Attraction\_Preference\_and\_Search,'' even though the latter already subsumes the former’s function. This results in unnecessary tool usage and inflated total cost.  
    \item \textbf{Extra Calls:} The model successfully reaches the goal state by Step~3 but continues to call additional tools instead of terminating the process. This behavior suggests a lack of awareness regarding task completion.
\end{itemize}

\subsubsection{Failure Calls}

Apart from redundant behavior, we observe two major categories of \textit{failure calls}, which directly lead to invalid tool invocations:

\paragraph{(1) Wrong Parameters.}  
These errors arise when the model specifies incorrect tool names or malformed parameter formats. For instance, a model may call a non-existent tool like ``Attraction\_Finish'' instead of the correct ``Attraction\_Finish\_from\_Step1\_3Steps.''  
This reflects inadequate grounding in the available tool schema and incomplete adherence to the prompt’s enumerated tool list (see Figure~\ref{fig:runtime-prompt}).

\paragraph{(2) Inaccessible Calls.}  
These represent the most severe failure mode. An inaccessible call occurs when the model invokes a tool whose input dependencies have not yet been satisfied.  
For example, after executing ``Search\_Location\_Candidates,'' a model may directly call ``Location\_Refinement\_Step2'' without completing ``Step~1'', which provides the required structured input.  
Such cases reveal a fundamental failure to reason about task dependencies, despite the explicit prompt instructions outlining the correct execution sequence.

As shown in Table~\ref{tab:error-analysis-statistics}, inaccessible calls dominate among all failure cases across models. This suggests that models struggle to maintain a consistent internal state of intermediate results or to map current progress to the corresponding valid action space.

\subsection{Summary and Implications}

Across all analyzed cases, the dominant failure modes, repeated, extra, wrong-parameter, and inaccessible calls, point to a shared underlying limitation: insufficient progress awareness.  
Models often fail to track which subgoals have been achieved and which inputs are currently available, leading to redundant or logically inconsistent tool invocations.  
This lack of situational grounding undermines cost sensitivity and prevents the models from performing truly cost-optimal planning, even when they possess the necessary cost-related reasoning capabilities.

\section{Data Annotation}
\label{app:data-annotation}

\begin{figure*}
    \centering
    \includegraphics[width=\linewidth]{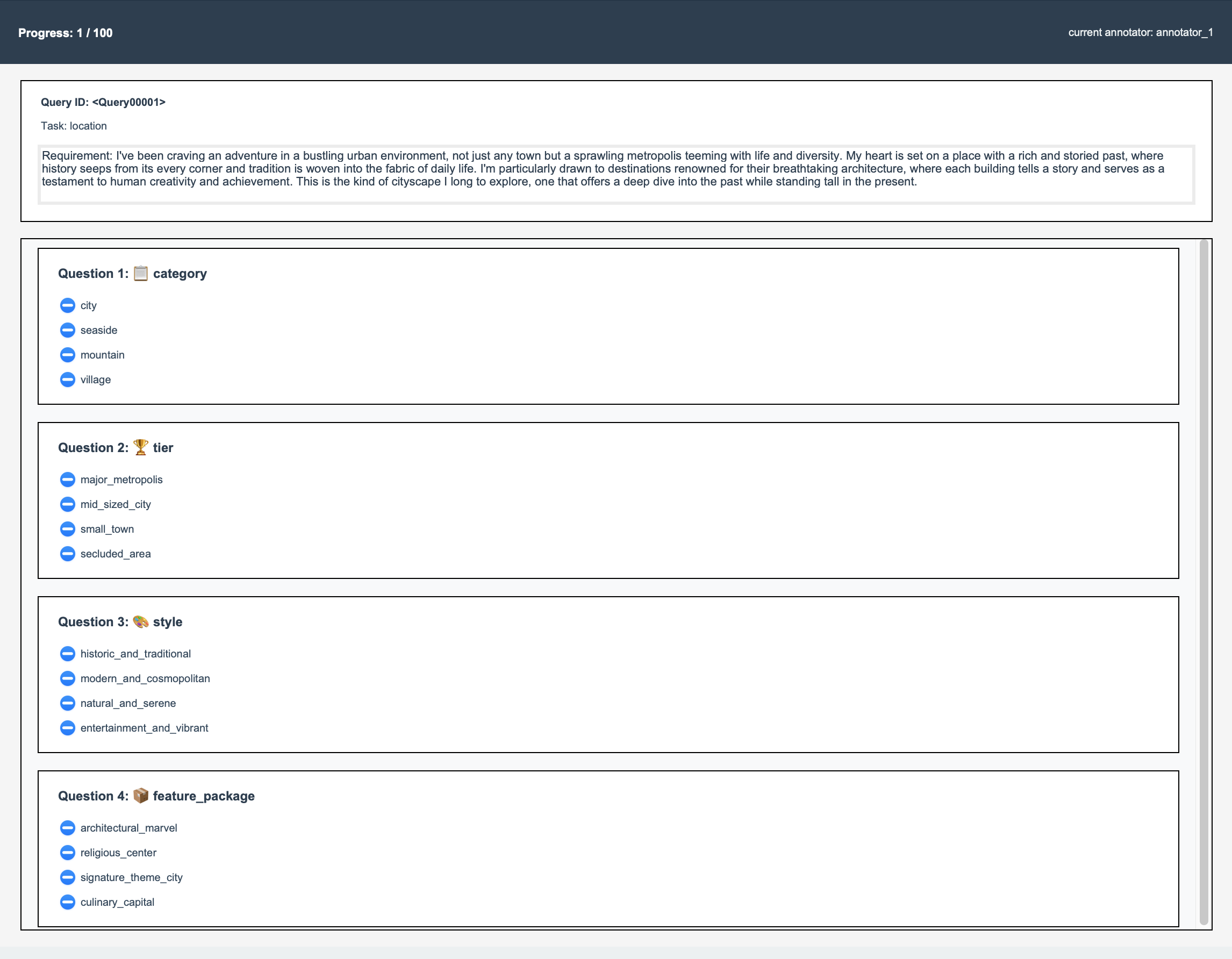}
    \caption{Annotation Screenshot for human annotators.}
    \label{fig:Annotation}
    \vspace{-0.1in}
\end{figure*}

\paragraph{Query Validation.} We enlisted three PhD-level researchers, all co-authors with expertise in NLP, to annotate a sample of 200 travel queries across six scenarios: accommodation, transportation, attractions, location, dining, and shopping.
A screenshot of the annotation interface is shown in Figure~\ref{fig:Annotation}.
The annotators achieved individual accuracy rates of 97\%, 95\%, and 98\% against the reference answers, confirming the high quality of the annotations. They also verified that the dataset contains no offensive language or personal information.

\paragraph{Coverage Rate Calculation.}
\label{app:coverage-rate-human-check}
For the calculation of the coverage rate described in Section~\ref{sec:analysis-for-static-setting}, we adopt the LLM-as-a-judge approach. To verify its reliability, we manually inspected 10 sampled cases for each setting and found the automatically extracted results to be fully consistent with the human annotations. Therefore, we are confident in the credibility of the reported results.

\begin{figure*}[h!]
\begin{tcolorbox}[
    colback=grey!3!white,
    colframe=black!30!white, 
    title={Prompts used in query construction stage},
    fonttitle=\bfseries,
    boxrule=0.5pt,
    arc=4pt,
    boxsep=5pt,
    left=6pt,
    right=6pt,
    top=6pt,
    bottom=6pt,
    coltitle=blue!50!black
]

\textbf{User preference validation}

You are a helpful assistant for validating commonsense conflicts in user queries.
Given a set of user requirements, determine whether there are any commonsense conflicts among them.
Apply strict checking: even minor inconsistencies should be marked as conflicts.
Your response must be either **conflict** or **no conflict**, nothing else. \\\\
Example:
User prompt: Task: Location search. User requirements: "
1. I want the Location category to be `city'. \\
2. I want the Location tier to be `secluded\_nature'. \\
3. I want the Location style to be `adventure'. \\
4. I want the Location features to include `nightlife\_central'.\\
Generated response: **conflict** \\

User prompt: \\\\

\textbf{Prompt construction}

You are a helpful assistant for generating queries. \\
Please generate a search query for a [task] task based on detailed user requirements. The user requirements will be comprised of four dimensions (Category requirement, Tier requirement, Style requirement, Feature package requirement). The query should be written as a long, self-contained user statement that clearly describes the user's needs and intentions. You should follow these rules: \\\\
1. The query clearly describe the user requirements without any possibilities of misunderstanding. For each requirement dimension, you should clearly distinguish the user required one from any other possible candidates in the generated query. Possible candidates are listed below: \\
Category requirement: [category\_candidates] \\
Tier requirement: [tier\_candidates] \\
Style requirement: [style\_candidates]  \\
Feature package requirement: [features\_candidates] \\
2. You should use human-like language to express the user requirements. That is to say, you shouldn't use the exact word to describe the user requirements. Instead, you should paraphrase and rephrase the requirements to imply the user needs in a natural way. For example: if the user has a 'luxury' requirement, then you could say something like 'money is totally not a concern, and I want an extravagant experience'. For some special cases (proper nouns), you can use the exact wording. \\
3. The query should be concise and to the point, avoiding unnecessary details or overly complex sentences. \\
4. All the information you could use is from the user preferences. The location and time information are just meaningless placeholders. \\\\
Please **DO NOT GENERATE ANYTHING OTHER THAN THE QUERY**.

\end{tcolorbox}
\caption{The prompts used in our query construction stage. All the words surrounded with ``[ ]'' would be replaced with real parameters in construction time.}
\label{fig:query-construction-prompt}
\end{figure*}

\begin{figure*}[h!]
\begin{tcolorbox}[
    colback=grey!3!white,
    colframe=black!30!white, 
    title={Prompts used in path extraction stage},
    fonttitle=\bfseries,
    boxrule=0.5pt,
    arc=4pt,
    boxsep=5pt,
    left=6pt,
    right=6pt,
    top=6pt,
    bottom=6pt,
    coltitle=blue!50!black
]

You are analyzing a planning text from a model's output. 

The text lists possible tool-calling paths with costs. Count the number of distinct paths explicitly enumerated (e.g., `Path 1', `1)', etc.). Ignore introductions or selected paths—only count the listed alternatives. You shouldn't judge if the paths are valid or not. Output a single integer to represent the number of distinct paths. If none, use 0. No extra text. 
\\\\
Planning text: [model\_plan]. Number of distinct paths: 

\end{tcolorbox}
\caption{The prompts used in our path extraction stage. All the word surrounded with ``[ ]'' would be replaced with real parameters in construction time.}
\label{fig:extract-converage-prompt}
\end{figure*}

\begin{figure*}[h!]
\begin{tcolorbox}[
    colback=grey!3!white,
    colframe=black!30!white, 
    title=An example user query,
    fonttitle=\bfseries,
    boxrule=0.5pt,
    arc=4pt,
    boxsep=5pt,
    left=6pt,
    right=6pt,
    top=6pt,
    bottom=6pt,
    coltitle=blue!50!black
]

\textbf{User Query}:

I've been craving an adventure in a bustling urban environment, not just any town but a sprawling metropolis teeming with life and diversity. My heart is set on a place with a rich and storied past, where history seeps from its every corner and tradition is woven into the fabric of daily life. I'm particularly drawn to destinations renowned for their breathtaking architecture, where each building tells a story and serves as a testament to human creativity and achievement. This is the kind of cityscape I long to explore, one that offers a deep dive into the past while standing tall in the present.
\\\\
\textbf{Corresponding user preference}:
\\
Location main category: city

Location scale tier: major\_metropolis

Location style priority: historical\_and\_traditional

Location feature package: architectural\_marvel

\end{tcolorbox}
\caption{An example user query in \cb{}.}
\label{fig:example-user-query}
\end{figure*}

\begin{figure*}[h!]
\begin{tcolorbox}[
    colback=grey!3!white,
    colframe=black!30!white, 
    title=An example of the tool used in our \CB,
    fonttitle=\bfseries,
    boxrule=0.5pt,
    arc=4pt,
    boxsep=5pt,
    left=6pt,
    right=6pt,
    top=6pt,
    bottom=6pt,
    coltitle=blue!50!black
]

\begin{lstlisting}
Name: Decide_Transportation_Preference 
Description: With this atomic tool, you describe a transportation by choosing its category, tier, style, and feature package according to the user requirements. The tool then gives you a unique label that represents your exact combination of transportation preferences. This tool has a cost of 20.06 units. The output type of this tool is TransportationPreference. 
Parameters: 
    - Type: object 
    - Properties: 
        - LocationPreference: 
            - Type: String 
            - Description: A LocationPreference object identifier representing user preferences for location selection 
        - TransportationCategory: 
            - Type: String
            - Enum: [
                flight,
                train,
                bus,
                car_rental
            ],
            - Description: A transportation category enumeration value specifying the main type/category for transportation selection. Available options: flight, train, bus, car_rental 
        - TransportationTier: 
            - Type: string
            - Enum: [luxury_class, business_class, standard_class, budget_class],
            - Description: A transportation tier enumeration value specifying the quality/price level for transportation selection. Available options: luxury_class, business_class, standard_class, budget_class.
        - TransportationStyle: 
            - Type: string
            - Enum: [speed_priority, comfort_priority, scenic_route, schedule_flexibility_priority],
            - Description: A transportation style enumeration value specifying the preferred style/approach for transportation selection. Available options: speed_priority, comfort_priority, scenic_route, schedule_flexibility_priority.
        - TransportationFeaturePackage: {
            - Type: String,
            - Enum: [onboard_connectivity_and_power, full_meal_and_beverage_service, special_luggage_allowance, lie_flat_or_sleeper_facility.
            ],
            - Description: A transportation feature package enumeration value specifying additional features/services for transportation selection. Available options: onboard_connectivity_and_power, full_meal_and_beverage_service, special_luggage_allowance, lie_flat_or_sleeper_facility.
    - Required: [LocationPreference, TransportationCategory, TransportationTier, TransportationStyle, TransportationFeaturePackage]
\end{lstlisting}

\end{tcolorbox}
\caption{An example tool schema for the tools we used in the \cb{}.}
\label{fig:sample-tools}
\end{figure*}

\begin{figure*}[h!]
\begin{tcolorbox}[
    colback=grey!3!white,
    colframe=black!30!white, 
    title=Inference Prompt Part 1,
    fonttitle=\bfseries,
    boxrule=0.5pt,
    arc=4pt,
    boxsep=5pt,
    left=6pt,
    right=6pt,
    top=6pt,
    bottom=6pt,
    coltitle=blue!50!black
]

You are an AI assistant for planning {task}-related schedules. 
\\\\
<Task description> \\
Your only objective is to obtain the required information (goal type: `TravelLocation', represented by a unique ID `<LocationCandidate\{Candidate\_ID\}>') by following the tool path with the **LOWEST TOTAL COST**. The task consists of 4 parts: Deciding Preference, Searching Candidates, Refining Options, Final Recommendation. In the refinement stage, you should take charge of filtering the Location candidates. You should refine the possible candidate set from these two dimensions: availability and seasonal suitability. Note that the order of the refinement steps is fixed as specified above, and using another order will result in incorrect behavior.\\
</Task description> 
\\\\
<Tool description> \\
1. **Tool Cost**. Each tool call has a predefined cost listed in the tool description. \\
2. **Tool Input and Output Types**. Each tool defines its input types through its parameters (the parameter name indicates the data type) and its output type in its description. \\
3. **Tool Dependencies**. Some tools depend on others through their input/output types. Carefully read each tool's input/output fields and description before calling the tool. \\
4. **Data types**. Each Tool has a list of input data types and an output data type. You should infer LocationCategory, LocationTier, LocationStyle, LocationFeaturePackage, TimeInfo from the user query. For other data types, you only obtain them when a certain tool explicitly returns them. The data types are specially designed, and using them incorrectly will result in incorrect behavior. \\
5. **Atomic vs Composite Tools**. The tools available could be categorized into atomic tools and composite tools, which is specified in the tool description. An atomic tool performs a single and inseparable operation. A composite tool chains multiple atomic tools in sequence and lists its component atomic tools in its description. The cost of a composite tool is specified in its description. Inputs/outputs of a composite tool follow the component chain. Despite being multi-step internally, it still counts as ONE tool call and must obey the one-tool-per-step rule. The cost of a composite tool might be higher or lower than the sum of its component atomic tools. \\
6. **Sample Atomic Tool Sequence**. For this task, the basic atomic tool calling sequence is: Decide\_Location\_Preference, Search\_Location\_Candidates, Location\_Refinement\_Step1, Select\_Final\_Location. You should replace some atomic tools with composite tools if that reduces cost. You must then compare all possible equivalent tool-calling paths and pick the one with the lowest total cost. \\
</Tool description> \\

\end{tcolorbox}
\caption{The first part of the prompt used to benchmark agents during runtime. The example shown corresponds to a task sequence of length 4. The filtering steps and demonstration examples are dynamically generated according to the task sequence length. This part is concatenated with Part 2 in Figure~\ref{fig:runtime-prompt-two} to form the complete prompt.}
\label{fig:runtime-prompt}
\end{figure*}

\begin{figure*}[h!]
\begin{tcolorbox}[
    colback=grey!3!white,
    colframe=black!30!white, 
    title=Inference Prompt Part 2,
    fonttitle=\bfseries,
    boxrule=0.5pt,
    arc=4pt,
    boxsep=5pt,
    left=6pt,
    right=6pt,
    top=6pt,
    bottom=6pt,
    coltitle=blue!50!black
]
<Expected workflow> \\
1. **Explain your reasoning.** Write out your plan clearly, showing how you’ll minimize cost. To ensure the optimality of your plan, you should list out all possible tool-calling paths, sum up the cost of each path, and then select the path with the lowest cost. \\
2. **Execute your plan.** Right after the explanation, invoke the required tool. Do not describe or print the tool call in text, just make the call directly. \\
3. **Adapt and continue.** You should always keep an eye on the environment. On every step of execution, you should always check if anything about the tool changes (e.g. cost, availability, etc.). If something goes wrong or changes, adapt and continue along the most cost-optimal path. \\
</Expected workflow> \\
\\
<Important rules> \\
- **Cost is the most important.** Your performance is evaluated solely based on the total cost of tool calls upon reaching the goal state. Always pick the cost-minimal tool path. If there are two paths with the same cost, you should pick the one with the least number of tool calls. \\
- **One tool per step.** You may only call one tool at a time and SHOULD NOT call multiple tools in one request. If you try to call multiple, only the first will count. \\
- **Exact parameters.** Use the provided values exactly as given (e.g., if ``<TimeInfo00000>'' is given, the ``TimeInfo'' parameter must be ``<TimeInfo00000>'', if ``<LocationPreference00000>'' is given, the ``LocationPreference'' parameter must be ``<LocationPreference00000>''). 
- **Final answer format.** Once you obtain the ``Candidate\_ID'' representing your goal type, stop calling tools immediately and return the answer in this exact format: ``<answer> <LocationCandidate\{Candidate\_ID\}> </answer>''. Only incorporate the ``<answer>'', ``</answer>'' tag when you want to provide the final answer. If you output the format, your conversation would be terminated. 
</Important rules> 
\\\\
<example> \\
Here is an example of how to plan your tool call paths in a cost-optimal way for your reference. You should adapt to the task and available tools instead of memorizing this example.\\
\\
Given that:
1. The basic atomic tool calling sequence is: A(Cost: Cost\_A), B(Cost: Cost\_B), C(Cost: Cost\_C), D(Cost: Cost\_D).
2. The available tools are: A(Cost: Cost\_A), B(Cost: Cost\_B), C(Cost: Cost\_C), D(Cost: Cost\_D), AB(Cost: Cost\_AB), BC(Cost: Cost\_BC), CD(Cost: Cost\_CD), ABC(Cost: Cost\_ABC), BCD(Cost: Cost\_BCD). Composite tools are those whose names contain at least two letters; each letter represents an atomic tool included within the composite, while their costs are not necessarily the sum of their component atomic tools (e.g., `AB' is equivalent in effect to performing A then B, but Cost\_AB may differ from Cost\_A + Cost\_B).\\
\\
Then you should list out all possible tool calling paths first:\\
{\small
<path> 1. A(Cost: Cost\_A) -> B(Cost: Cost\_B) -> C(Cost: Cost\_C) -> D(Cost: Cost\_D). Total Cost: Cost\_A + Cost\_B + Cost\_C + Cost\_D.</path>
<path> 2. AB(Cost: Cost\_AB) -> C(Cost: Cost\_C) -> D(Cost: Cost\_D). Total Cost: Cost\_AB + Cost\_C + Cost\_D.</path>
<path> 3. A(Cost: Cost\_A) -> BC(Cost: Cost\_BC) -> D(Cost: Cost\_D). Total Cost: Cost\_A + Cost\_BC + Cost\_D.</path>
<path> 4. A(Cost: Cost\_A) -> B(Cost: Cost\_B) -> CD(Cost: Cost\_CD). Total Cost: Cost\_A + Cost\_B + Cost\_CD.</path>
<path> 5. AB(Cost: Cost\_AB) -> CD(Cost: Cost\_CD). Total Cost: Cost\_AB + Cost\_CD.</path>
<path> 6. ABC(Cost: Cost\_ABC) -> D(Cost: Cost\_D). Total Cost: Cost\_ABC + Cost\_D.</path>
<path> 7. A(Cost: Cost\_A) -> BCD(Cost: Cost\_BCD). Total Cost: Cost\_A + Cost\_BCD.</path>}\\
At last, you should select and execute the path with the lowest total cost.
</example>

\end{tcolorbox}
\caption{The second part for agent runtime prompt. }
\label{fig:runtime-prompt-two}
\end{figure*}

\begin{table*}[t]
    \centering
    \begin{tabularx}{\textwidth}{l|X}
        \toprule
        Dimension & Values \\
        \midrule
        Category & flight, train, bus, car rental \\
        Tier & luxury class, business class, standard class, budget class \\
        Style & speed priority, comfort priority, scenic route, schedule flexibility priority \\
        Feature Package & onboard connectivity and power, full meal and beverage service, special luggage allowance, lie flat or sleeper facility \\
        \bottomrule
	\end{tabularx}
    \caption{The test set feature list for task transportation.}
    \label{tab:feature-list-transportation}
\end{table*}

\end{document}